\pdfoutput=1
\documentclass[11pt]{article}
\usepackage{acl}
\usepackage{times}
\usepackage{latexsym}
\usepackage[utf8]{inputenc}
\usepackage{microtype}
\usepackage{hyperref}
\usepackage{url}
\usepackage{float}
\usepackage{mathtools}
\usepackage[disable]{todonotes}
\usepackage{booktabs}
\usepackage{graphicx}
\usepackage{xspace,mfirstuc,tabulary}

\newcommand{\bftab}{\fontseries{b}\selectfont}
\newcommand{\newbftab}[1]{\fontseries{b}\selectfont \underline{#1}}
\newcommand{\semeval}{{SemEval}\xspace}
\newcommand{\deisear}{{deISEAR}\xspace}
\newcommand{\sab}{{SAB}\xspace}
\newcommand{\sbtenk}{{sb10k}\xspace}
\newcommand{\amazonreviews}{{Amazon}\xspace}
\newcommand{\unified}{{Unified}\xspace}
\newcommand{\dailydialog}{{DailyDialog}\xspace}
\newcommand{\crowdflower}{{CrowdFlower}\xspace}
\newcommand{\tec}{{TEC}\xspace}
\newcommand{\talesemotion}{{Tales}\xspace}
\newcommand{\isear}{{ISEAR}\xspace}
\newcommand{\emoint}{{Emoint}\xspace}
\newcommand{\electoraltweets}{{ElectoralTweets}\xspace}
\newcommand{\groundedemotions}{{GroundedEmotions}\xspace}
\newcommand{\emotioncause}{{EmotionCause}\xspace}
\newcommand{\yahoo}{{Yahoo}\xspace}
\newcommand{\subj}{{SUBJ}\xspace}
\newcommand{\cola}{{COLA}\xspace}
\newcommand{\trec}{{TREC}\xspace}
\newcommand{\yelp}{{Yelp}\xspace}
\newcommand{\agnews}{{AG News}\xspace}
\newcommand{\imdb}{{IMDB}\xspace}
\newcommand{\headqa}{HeadQA\xspace}
\newcommand{\gnad}{GNAD\xspace}

\newcommand{\roberta}{\texttt{RoBERTa}\xspace}
\newcommand{\robertanatcat}{RoBERTa-NatCat\xspace}

\newcommand{\mpnet}{MPNET\xspace}
\newcommand{\snli}{SNLI\xspace}
\newcommand{\mnli}{MNLI\xspace}
\newcommand{\bitfit}{BitFit\xspace}

\newif\ifspacehacks
\spacehacksfalse
\ifspacehacks
\newcommand{\mysubsection}[1]{\paragraph{#1}}
\else
\newcommand{\mysubsection}[1]{\subsection{#1}}
\fi

\title{Few-Shot Learning with Siamese Networks and Label Tuning}

\author{
Thomas M{\"u}ller \and
Guillermo Pérez-Torró \and
Marc Franco-Salvador\\
Symanto Research, Valencia, Spain\\
\url{https://www.symanto.com}\\
\texttt{\{thomas.mueller,guillermo.perez,marc.franco\}@symanto.com}
}

\begin{document}
\maketitle
\begin{abstract}
We study the problem of building text classifiers with little or no training data, commonly known as zero and few-shot text classification.
In recent years, an approach based on neural textual entailment models has been found to give strong results on a diverse range of tasks.
In this work, we show that with proper pre-training, Siamese Networks that embed texts and labels offer a competitive alternative.
These models allow for a large reduction in inference cost: constant in the number of labels rather than linear.
Furthermore, we introduce label tuning, a simple and computationally efficient approach that allows to adapt the models in a few-shot setup by only changing the label embeddings.
While giving lower performance than model fine-tuning, this approach has the architectural advantage that a single encoder can be shared by many different tasks.
\end{abstract}

\section{Introduction}

% https://docs.google.com/drawings/d/1xz4sgpELU2Pc0o1IoPXIMg1DkKKEi0k4wgUDtNkVADQ
\begin{figure*}
    \centering
    \includegraphics[width=400pt]{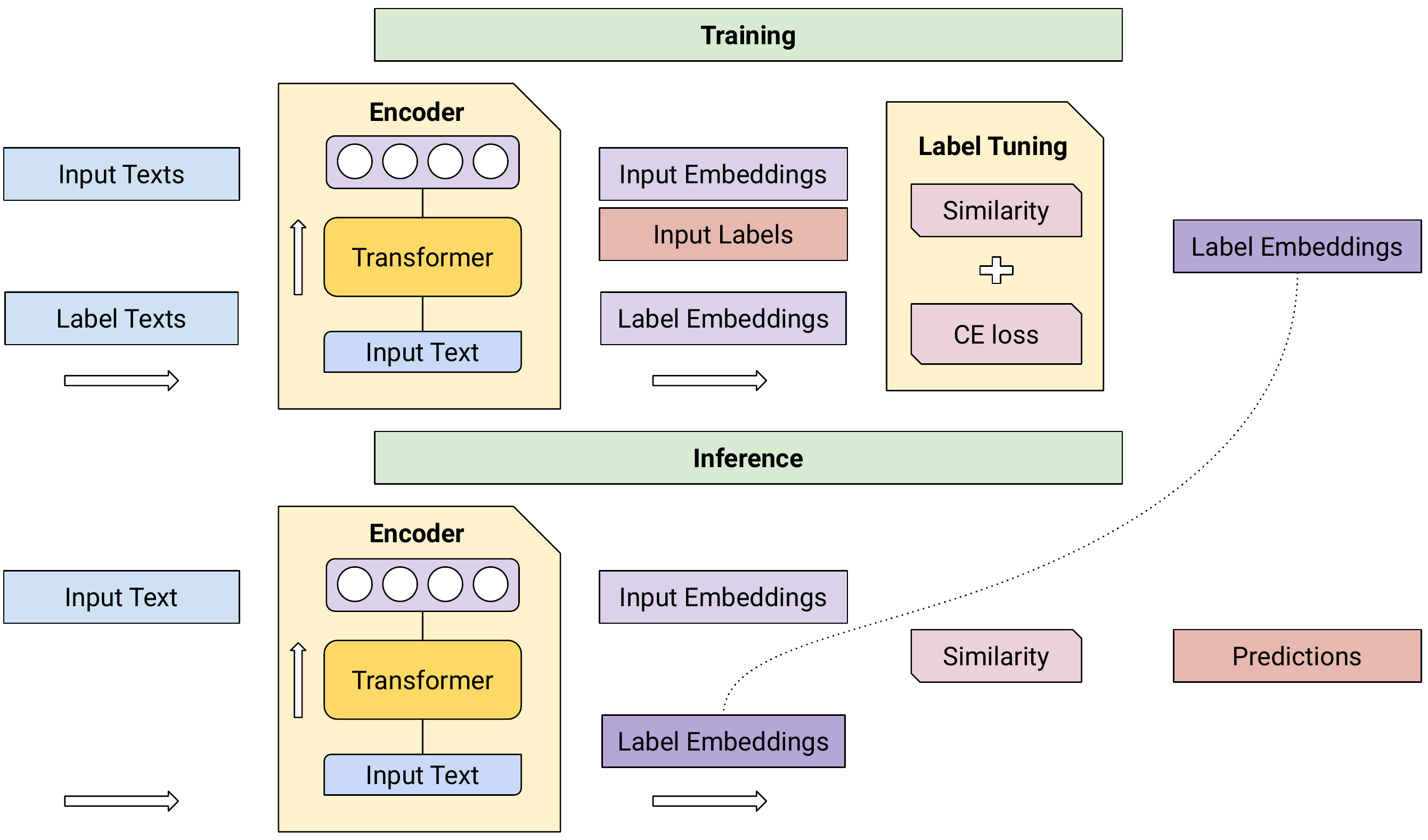}
    \caption{Overview of training and inference with Label Tuning (LT). At training time, input and label texts (hypotheses) are processed by the encoder. LT then tunes the labels using a cross entropy (CE) loss.
    At inference time, the input text is passed through the same encoder. The tuned label embeddings and a similarity function are then used to score each label.
    The encoder remains unchanged and can be shared between multiple tasks.}
    \label{fig:siamese_network}
\end{figure*}

Few-shot learning is the problem of learning classifiers with only a few training examples.
Zero-shot learning \citep{larochelle2008}, also known as dataless classification \citep{Chang2008ImportanceOS},
is the extreme case, in which no labeled data is used.
For text data, this is usually accomplished by representing the labels of the task in a textual form, which can either be the name of the label or a concise textual description.

In recent years, there has been a surge in zero-shot and few-shot approaches to text classification.
One approach \citep{yin-etal-2019-benchmarking,yin-etal-2020-universal,halder-etal-2020-task,Wang2021EntailmentAF}
makes use of entailment models.
Textual entailment \citep{10.1007/11736790_9}, also known as natural language inference (NLI) \citep{bowman-etal-2015-large}, is the problem of predicting whether a textual premise implies a textual hypothesis in a logical sense. 
For example, \emph{Emma loves apples} implies that \emph{Emma likes apples}. 

The entailment approach for text classification sets the input text as the premise and the text representing the label as the hypothesis. 
A NLI model is applied to each input pair and the entailment probability is used to identify the best matching label.

In this paper, we investigate an alternative based on Siamese Networks (SN) \citep{10.5555/2987189.2987282}, also known as dual encoders.
These models embed both input and label texts into a common vector space. 
The similarity of the two items can then be computed using a similarity function such as the dot product. 
The advantage is that input and label text are encoded independently, which means that the label embeddings can be pre-computed.
Therefore, at inference time, only a single call to the model per input is needed. 
In contrast, the models typically applied in the entailment approach are Cross Attention (CA) models which need to be executed for every combination of text and label. 
On the other hand, they allow for interaction between the tokens of label and input, so that in theory they should be superior in classification accuracy. 
However, in this work we show that in practice, the difference in quality is small.

Both CA and SNs also support the few-shot learning setup by fine-tuning the models on a small number of labeled examples.
This is usually done by updating all parameters of the model, which in turn makes it impossible to share the models between different tasks. 
In this work, we show that when using a SN, one can decide to only fine-tune the label embeddings. 
We call this Label Tuning (LT).
With LT the encoder can be shared between different tasks, which greatly eases the deployment of this approach in a production setup. 
LT comes with a certain drop in quality, but this drop can be compensated by using a variant of knowledge distillation \citep{Hinton2015DistillingTK}.

Our contributions are as follows:
We perform a large study on a diverse set of tasks showing that CA models and SN yield similar performance for both zero-shot and few-shot text classification.
In contrast to most prior work, we also show that these results can also be achieved for languages other than English.
We compare the hypothesis patterns commonly used in the literature and using the plain label name (identity hypothesis) and find that on
average there is no significant difference in performance.
Finally, we present LT as an alternative to full fine-tuning that allows using the same model for many tasks and thus greatly increases the scalability of the method.
We will release the code\footnote{\url{https://tinyurl.com/label-tuning}} and trained models used in our experiments.

\section{Methodology}

% \begin{figure}
%     \centering
%     \includegraphics[width=200pt]{figures/siamese_network_architecture.pdf}
%     \caption{A Siamese network applied to text classification. The same transformer model is used to map the input and label text to a fixed-length embedding.
%     A similarity function (dot product or cosine similarity) computes the scalar similarity.}
%     \label{fig:siamese_network}
% \end{figure}

Figure \ref{fig:siamese_network} explains the overall system.
We follow \citet{reimers-2019-sentence-bert} and apply symmetric Siamese Networks that embed both input texts using a single encoder.
The encoder consists of a transformer \citep{NIPS2017_3f5ee243} that produces contextual token embeddings and a mean pooler that combines the token embeddings into a single text embedding. 
We use the dot product as the similarity function.
We experimented with cosine similarity but did not find it to yield significantly better results.

As discussed, we can directly apply this model to zero-shot text classification by embedding the input text and a textual representation of the label.
For the label representation we experiment with a plain verbalization of the label, or identity hypothesis% %(analogous to \textit{null prompt} \citet{logan2021})
, as well as the hypotheses or prompts used in the related work. %\citep{yin-etal-2019-benchmarking,gao2020,Wang2021EntailmentAF}.

\mysubsection{Fine-Tuning}
\label{sec:finetuning}
In the case of few-shot learning, we need to adapt the model based on a small set of examples.
In gradient-based few-shot learning we attempt to improve the similarity scores for a small set of labeled examples.
Conceptually, we want to increase the similarity between every text and its correct label and decrease the similarity for every other label.
As the objective we use the so called \emph{batch softmax} \citep{HendersonASSLGK17}:
\[
    \small
    \mathcal{J} = - \frac{1}{B} \sum_{i=1}^{B} \left[ S(x_i, y_i) - \log \sum_{j=1}^{B} e^{S(x_i, y_j)}  \right]
\]
Where $B$ is the batch size and $S(x,y) = f(x) \cdot f(y)$ the similarity between input $x$ and label text $y$ under the current model $f$.
All other elements of the batch are used as \emph{in-batch negatives}.
To this end, we construct the batches so that every batch contains exactly one example of each label.
Note that this is similar to a typical softmax classification objective.
The only difference is that $f(y_i)$ is computed during the forward pass and not as a simple parameter look-up.

\mysubsection{Label Tuning}
Regular fine-tuning has the drawback of requiring to update the weights of the complete network.
This results in slow training and large memory requirements for every new task, which in turn makes it challenging to deploy new models at scale.
As an alternative, we introduce \textbf{label tuning}, which does not change the weights of the encoder.  
The main idea is to first pre-compute label embeddings for each class and later tune them using a small set of labeled examples. 
Formally, we have a training set containing $N$ pairs of an input text $x_i$ and its reference label index $z_i$.
We pre-compute a matrix of the embedded input texts and embedded labels, $X {\in} \mathcal{R}^{N\times d}$ and $Y {\in} \mathcal{R}^{K\times d}$, respectively.
$d$ is the embedding dimension and $K$ the size of the label set. 
We now define the score for every input and label combination as $S = X\times Y^T$ ($S{\in} \mathcal{R}^{N\times K}$) and  tune it using cross entropy:
\[
    \mathcal{J'} = - \frac{1}{N} \sum_{i=1}^{N} \left[ S_{i,z_i} - \log \sum_{j=1}^{K} e^{S_{i,j}}  \right]
\]
To avoid overfitting, we add a regularizer that penalizes moving too far from the initial label embeddings $Y_0$ as $\lVert Y_0-Y\rVert_F$,
where $\lVert.\rVert_F$ is the Frobenius norm.\footnote{\url{https://en.wikipedia.org/wiki/Matrix_norm\#Frobenius\_norm}}
Additionally, we also implement a version of dropout by masking some of the entries in the 
label embedding matrix at each gradient step. 
To this end, we sample a random vector $\vec r$ of dimension $d$ whose components are $0$ with probability \emph{dropout} and $1$ otherwise. 
We then multiply this vector component-wise with each row in the label embedding matrix $Y$.
The dropout rate and the strength of the regularizer are two hyper-parameters of the method. 
The other hyper-parameters are the learning rate for the stochastic gradient descent as well as the number of steps. 
Following \citet{logan2021}, we tune them using 4-fold cross-validation on the few-shot training set.
Note that the only information to be stored for each tuned model are the $d$-dimensional label embeddings.

\mysubsection{Knowledge Distillation}
As mentioned, label tuning produces less accurate models than real fine-tuning.
We find that this can be compensated by a form of knowledge distillation \citep{Hinton2015DistillingTK}.
We first train a normal fine-tuned model and use that to produce label distributions for a set of unlabeled examples. 
Later, this silver set is used to train the new label embeddings for the untuned model. 
This increases the training cost of the approach and adds an additional requirement of unlabeled data but keeps the advantages that at inference time we can
share one model across multiple tasks.

\section{Related Work}

Pre-trained Language Models (LMs) have been proved to encode knowledge that, with task-specific guidance, can solve natural language understanding tasks \citep{petroni-etal-2019-language}.
Leveraging that, \citet{le-scao-rush-2021-many} quantified a reduction in the need of labeled data of hundreds of instances with respect to traditional fine-tuning approaches \citep{devlin-etal-2019-bert,DBLP:journals/corr/abs-1907-11692}.
This has led to quality improvements in zero and few-shot learning.

% The transformer architecture \citep{NIPS2017_3f5ee243} and pre-training on large text corpora \citep{devlin-etal-2019-bert,DBLP:journals/corr/abs-1907-11692} have 
% led to
% improvements on many supervised text classification tasks.
% However, fine-tuning these models still requires considerable amounts of human-labeled data. 

% %Aiming to find less data hungry approaches, the community has focused intensely on new methodologies.
% %These try to mimic the human capability to categorize something with few examples, or even anyone, just understanding the meaning of the task and the appropriate labeling. 
% New approaches propose to reduce data needs in classification tasks \citep{le-scao-rush-2021-many} by explicitly incorporating tasks descriptions and making use of the knowledge encoded in pre-trained Language Models (LMs) \citep{petroni-etal-2019-language}.
% This has led to quality improvements in zero and few-shot learning.

\paragraph{Semantic Similarity methods}
%Some of the firsts attempts to implement a zero-shot approach in natural language processing were carried out by \citet{Gabrilovich2007ComputingSR} and \citet{Chang2008ImportanceOS}. 
\citet{Gabrilovich2007ComputingSR} and \citet{Chang2008ImportanceOS} use the explicit meaning of the label names to compute the similarity with the input text.
Prototypical Networks \citep{snell2017prototypical} create class prototypes by averaging embedded support examples and minimizing a distance metric to them for classification of input examples.
The class prototypes are similar to our label embeddings but we initialize them from the hypotheses and only tune the embeddings instead of the entire encoder.
% \citet{sui-etal-2021-knowledge} use PNs for few-shot text classification although prototype classes are not learnable parameters as in our proposed \textbf{label tuning} method.
Recent advances in pre-trained LMs and their application to semantic textual similarity tasks, such as Sentence-BERT \citep{reimers-2019-sentence-bert}, 
have shown a new opportunity to increase the quality of these methods and set the stage for this work.
\citet{baldini-soares-etal-2019-matching} use Siamese Networks applied to a few-shot relation extraction (RelEx) task.
Their architecture and similarity loss is similar to ours, but they update all encoder parameters when performing fine-tuning.
% They differ from us in two aspects: (i) pre-training the encoder with a novel setup specific for RelEx tasks, \emph{Matching the Blanks}; (ii) updating all encoder parameters when performing task-specific fine-tuning.
\citet{chu2020} employ a technique called \emph{unsupervised label-refinement} (LR).
They incorporate a modified \emph{k-means} clustering algorithm for refining the outputs of cross attention and Siamese Networks. 
We incorporate LR into our experiments and extend the analysis of their work. 
We evaluate it against more extensive and diverse benchmarks. 
In addition, we show that pre-training few-shot learners on their proposed textual similarity task NatCat underperforms pre-training on NLI datsets.

\paragraph{Prompt-based methods}
% \emph{Prompt-based learning} \citep{liu2021pretrain} is based on large pre-trained LMs.
GPT-3  \citep{NEURIPS2020_1457c0d6}, a 175 billion parameter LM, has been shown to give good quality on few-shot learning tasks.
Pattern-Exploiting Training (PET) \citep{schick-schutze-2021-exploiting} 
%and its successive variants iterative-PET (iPET) \citep{schick-schutze-2021-just} 
%and ADAPET \citep{Tam2021ImprovingAS} 
is a more computational and memory efficient alternative.
It is based on ensembles of smaller masked language models (MLMs) and was found to give few-shot results similar to GPT-3.
\citet{logan2021} reduced the complexity of finding optimal templates in PET by using \emph{null-prompts} and achieved competitive performance.
They incorporated \bitfit \citep{BenZaken202} and thus reached comparable accuracy fine-tuning only 0.1\% of the parameters of the LMs.
\citet{hambardzumyan-etal-2021-warp} present a contemporary approach with a similar idea to \textbf{label tuning}. 
As in our work, they use label embeddings initialized as the verbalization of the label names.
These task-specific embeddings, along with additional ones that are inserted into the input sequence, are the only learnable parameters during model training. 
They optimize a cross entropy loss between the label embeddings and the output head of a MLM.
The major difference is that they employ a prompt-based approach while our method relies on embedding models.

\paragraph{Entailment methods}
\label{lr:entailemnt}
The entailment approach \citep{yin-etal-2019-benchmarking, halder-etal-2020-task}
uses the label description to reformulate text classification as textual entailment. 
The model predicts the entailment probability of every label description .
% Like PET, it is based on smaller LMs and shows equally promising results.
\citet{Wang2021EntailmentAF} report results outperforming 
LM-BFF \citep{gao2020}, an approach similar to PET.

\paragraph{True Few-Shot Learning Setting} 
\citet{perez2021true} argue that for \textit{true few-shot learning}, one should not tune parameters on large validation sets or
use parameters or prompts that might have been tuned by others.
We follow their recommendation and rely on default parameters
and some hyper-parameters and prompts recommended by \citet{Wang2021EntailmentAF}, which according to the authors, were not tuned on the few-shot datasets.
For label tuning, we follow \citet{logan2021} and tune parameters with cross-validation on the few-shot training set.

\section{Experimental Setup}

In this section we introduce the baselines and datasets used throughout experiments.

\begin{table*}[t]
\centering
\resizebox{0.9\textwidth}{!}{%
\begin{tabular}{lrcrrrr}
\toprule
          name & task & lang. &  train &    test &  labels &  token length \\
\midrule
       \gnad \cite{Gnad} & topic    & de &   9,245 &    1,028 &       9 &         279 \\
       \agnews \cite{AgNews}   & &en &  120,000 &    7,600 &       4 &          37 \\
       \multicolumn{2}{l}{\headqa \cite{vilares-gomez-rodriguez-2019-head}} & es &   4,023 &    2,742 &       6 &          15 \\
         \yahoo \cite{NIPS2015_250cf8b5}   && en & 1,360,000 &  100,000 &      10 &          71 \\
\midrule
   \amazonreviews Reviews \cite{DBLP:journals/corr/abs-2010-02573} & reviews & de, en, es &   205,000 &    5,000 &       5 &          25-29 \\
          \imdb \cite{maas-EtAl:2011:ACL-HLT2011}  &&  en  & 25,000 &   25,000 &       2 &         173 \\
     \yelp full \cite{NIPS2015_250cf8b5}  && en &  650,000 &   50,000 &       5 &          99 \\
      \yelp polarity \cite{NIPS2015_250cf8b5}  && en &  560,000 &   38,000 &       2 &          97 \\
\midrule
\sab \cite{10.1007/978-3-319-66429-3_68} & sentiment &  es &   3,979 &     459 &       3 &          13 \\
\semeval \cite{SemEval:2016:task4}    && en &    9,834 &   20,632 &       3 &          20 \\
\sbtenk \cite{cieliebak-etal-2017-twitter} && de &    8,955 &     994 &       3 &          11 \\
\midrule
\unified \cite{Bostan2018}  & emotions & en &   42,145 &   15,689 &      10 &          15 \\
\deisear \cite{troiano-etal-2019-crowdsourcing}   && de &     643 &     340 &       7 &           9 \\
\midrule
\cola \cite{warstadt-etal-2019-neural}  & acceptability & en &    8,551 &    1,043 &       2 &           7 \\
\subj \cite{pang-lee-2004-sentimental}  & subjectivity & en &   8,019 &    1,981 &       2 &          22 \\
\trec \cite{li-roth-2002-learning}  & entity type & en &    5,452 &     500 &       6 &          10 \\
\bottomrule
\end{tabular}
}
\caption{Overview of the evaluated datasets. Token length is the median value.}
\label{tab:dataset_stats}
\end{table*}

\begin{table*}[t]
\centering
\resizebox{\textwidth}{!}{%
\begin{tabular}{lrllllllllllll}
\toprule
           name &    n &               \yahoo &              \agnews &             \unified &                 \cola &                \subj &                \trec &                \imdb &             \semeval &            \yelp pol &           \yelp full &       \amazonreviews &          Mean \\
\midrule
         random &    0 &                 10.0 &                 25.0 &                 10.0 &                  \bftab 50.0 &                 50.0 &                 16.7 &                 50.0 &                 33.3 &                 50.0 &                 20.0 &                 20.0 &          30.5 \\
       W2V (IH) &    0 &         44.8$_{0.2}$ &         59.1$_{0.5}$ &         10.1$_{0.3}$ &          46.9$_{1.7}$ &         37.1$_{0.7}$ &         17.6$_{1.4}$ &         71.0$_{0.3}$ &  \newbftab{46.8}$_{0.3}$ &         65.9$_{0.2}$ &         14.8$_{0.1}$ &         17.8$_{0.4}$ &  39.3$_{0.7}$ \\
 \robertanatcat &    0 &         50.0$_{0.2}$ &         49.8$_{0.6}$ &          7.9$_{0.3}$ &          35.5$_{1.5}$ &         44.3$_{0.9}$ &         18.6$_{1.1}$ &         45.6$_{0.3}$ &         36.6$_{0.3}$ &         49.8$_{0.2}$ &         11.1$_{0.1}$ &         11.2$_{0.4}$ & 32.8$_{0.7}$ \\
 \robertanatcat (IH) &    0 &         37.3$_{0.2}$ &         62.6$_{0.5}$ &         15.2$_{0.3}$ &          42.3$_{1.4}$ &         40.4$_{1.0}$ &         22.2$_{1.2}$ &         39.9$_{0.2}$ &         30.9$_{0.3}$ &         47.7$_{0.2}$ &         17.5$_{0.1}$ &         17.5$_{0.5}$ &  33.9$_{0.7}$ \\
     mpnet (CA) &    0 &         51.8$_{0.1}$ &         60.5$_{0.6}$ &  \newbftab{23.3}$_{0.4}$ &          47.0$_{1.4}$ &         41.0$_{0.9}$ &         19.8$_{1.6}$ &  \newbftab{87.5}$_{0.2}$ &         37.4$_{0.3}$ &  \newbftab{88.4}$_{0.2}$ &  \newbftab{36.7}$_{0.2}$ &         25.6$_{0.6}$ &  47.2$_{0.8}$ \\
  mpnet (CA-IH) &    0 &         46.3$_{0.2}$ &         56.3$_{0.5}$ &         22.2$_{0.4}$ &          47.7$_{1.5}$ &  \newbftab{55.7}$_{1.1}$ &         20.2$_{1.5}$ &         83.5$_{0.2}$ &         38.8$_{0.2}$ &         83.4$_{0.2}$ &         36.1$_{0.2}$ &         33.4$_{0.6}$ &  47.6$_{0.8}$ \\
     mpnet (SN) &    0 &  \newbftab{53.9}$_{0.1}$ &         62.5$_{0.5}$ &         21.6$_{0.3}$ &          46.0$_{1.5}$ &         42.0$_{0.8}$ &         31.5$_{1.4}$ &         73.8$_{0.2}$ &         46.7$_{0.3}$ &         78.6$_{0.2}$ &         26.1$_{0.2}$ &  \newbftab{40.6}$_{0.6}$ &  47.6$_{0.7}$ \\
  mpnet (SN-IH) &    0 &         51.4$_{0.1}$ &  \newbftab{64.2}$_{0.6}$ &         21.2$_{0.3}$ &          46.0$_{1.6}$ &         54.0$_{1.0}$ &  \newbftab{32.1}$_{1.7}$ &         69.6$_{0.3}$ &         41.5$_{0.3}$ &         83.6$_{0.2}$ &         34.3$_{0.2}$ &         37.4$_{0.7}$ &  48.7$_{0.8}$ \\
\midrule
       Char-SVM &    8 &         29.3$_{1.6}$ &         54.3$_{2.5}$ &         12.2$_{1.1}$ &          45.6$_{1.8}$ &         64.9$_{3.9}$ &         39.5$_{3.9}$ &         57.1$_{3.5}$ &         33.6$_{1.1}$ &         56.7$_{5.4}$ &         29.2$_{1.8}$ &         30.0$_{1.6}$ &  41.1$_{2.9}$ \\
     mpnet (CA) &    8 &  \underline{58.3}$_{2.8}$ &  \underline{80.6}$_{2.9}$ &  \underline{23.6}$_{1.1}$ &   \newbftab{50.4}$_{2.1}$ &  \underline{75.2}$_{5.0}$ &  \underline{66.4}$_{6.0}$ &  \newbftab{88.4}$_{0.9}$ &  \newbftab{59.5}$_{1.3}$ &  \newbftab{90.3}$_{1.9}$ &  \newbftab{50.9}$_{2.1}$ &  \newbftab{47.7}$_{1.3}$ &  62.8$_{2.9}$ \\
  mpnet (CA-IH) &    8 &  \underline{59.2}$_{2.6}$ &  \underline{83.1}$_{1.7}$ &  \underline{23.0}$_{2.2}$ &   \underline{48.4}$_{2.2}$ &  \underline{74.6}$_{5.3}$ &  \newbftab{68.7}$_{7.7}$ &  \underline{87.2}$_{0.8}$ &  \underline{58.2}$_{1.0}$ &  \underline{88.9}$_{3.8}$ &  \underline{49.3}$_{2.4}$ &  \underline{47.3}$_{1.7}$ &  62.5$_{3.5}$ \\
     mpnet (SN) &    8 &  \newbftab{62.0}$_{0.4}$ &  \underline{84.2}$_{1.5}$ &  \newbftab{24.8}$_{1.3}$ &   \underline{49.6}$_{1.8}$ &  \underline{79.6}$_{5.4}$ &  \underline{62.8}$_{6.4}$ &         76.4$_{1.6}$ &  \underline{58.7}$_{2.4}$ &         84.8$_{1.8}$ &         44.7$_{2.0}$ &  \underline{46.9}$_{1.7}$ &  61.3$_{3.0}$ \\
  mpnet (SN-IH) &    8 &  \underline{61.0}$_{0.9}$ &  \newbftab{84.4}$_{1.2}$ &  \underline{24.6}$_{1.1}$ &          46.3$_{2.7}$ &  \newbftab{80.5}$_{5.0}$ &  \underline{58.5}$_{2.4}$ &         76.1$_{1.9}$ &  \underline{57.0}$_{3.2}$ &         86.2$_{0.4}$ &         43.5$_{1.8}$ &  \underline{46.0}$_{1.8}$ &  60.4$_{2.4}$ \\
\midrule
       Char-SVM &   64 &         49.0$_{0.5}$ &         76.6$_{0.6}$ &         17.3$_{0.4}$ &          48.5$_{1.6}$ &         79.6$_{1.2}$ &         60.4$_{2.2}$ &         70.9$_{1.5}$ &         39.0$_{0.8}$ &         77.3$_{2.5}$ &         41.8$_{0.4}$ &         43.5$_{0.8}$ &  54.9$_{1.3}$ \\
     mpnet (CA) &   64 &  \underline{66.5}$_{0.9}$ &  \newbftab{87.9}$_{0.9}$ &  \underline{28.1}$_{1.3}$ &   \underline{54.2}$_{0.8}$ &  \underline{91.6}$_{1.4}$ &  \underline{87.0}$_{1.9}$ &  \newbftab{90.7}$_{1.0}$ &  \underline{62.0}$_{2.4}$ &  \newbftab{93.5}$_{0.4}$ &  \newbftab{57.0}$_{0.4}$ &  \newbftab{54.1}$_{1.5}$ &  70.2$_{1.3}$ \\
  mpnet (CA-IH) &   64 &         65.8$_{0.4}$ &  \underline{87.4}$_{1.0}$ &         26.4$_{0.6}$ &          51.3$_{2.2}$ &  \underline{92.5}$_{0.5}$ &  \underline{85.0}$_{2.1}$ &         89.3$_{0.5}$ &  \newbftab{62.6}$_{1.5}$ &         92.7$_{0.4}$ &         56.1$_{0.6}$ &  \newbftab{54.1}$_{1.3}$ &  69.4$_{1.2}$ \\
     mpnet (SN) &   64 &  \newbftab{66.6}$_{0.4}$ &  \underline{87.7}$_{1.0}$ &  \newbftab{29.3}$_{0.3}$ &   \newbftab{56.6}$_{1.8}$ &  \underline{92.0}$_{1.0}$ &  \underline{87.7}$_{1.9}$ &         79.7$_{1.4}$ &  \underline{61.9}$_{1.2}$ &         88.7$_{0.4}$ &         50.8$_{0.9}$ &  \newbftab{54.1}$_{1.4}$ &  68.6$_{1.2}$ \\
  mpnet (SN-IH) &   64 &  \underline{66.5}$_{0.4}$ &  \underline{87.3}$_{1.2}$ &  \newbftab{29.3}$_{0.5}$ &  \underline{46.5}$_{11.0}$ &  \newbftab{92.7}$_{0.3}$ &  \newbftab{87.5}$_{3.1}$ &         79.7$_{1.6}$ &  \underline{61.5}$_{1.7}$ &         88.1$_{0.2}$ &         50.7$_{0.8}$ &  \underline{54.0}$_{1.7}$ &  67.6$_{3.6}$ \\
\midrule
       Char-SVM &  512 &         59.6$_{0.2}$ &         85.8$_{0.3}$ &         23.0$_{0.4}$ &          51.2$_{1.1}$ &         87.0$_{0.6}$ &         87.5$_{0.7}$ &         82.8$_{0.5}$ &         46.0$_{0.5}$ &         87.1$_{0.2}$ &         49.3$_{0.3}$ &         50.4$_{0.4}$ &  64.5$_{0.5}$ \\
     mpnet (CA) &  512 &         67.1$_{0.7}$ &  \underline{90.2}$_{0.4}$ &  \underline{32.4}$_{1.2}$ &          68.5$_{2.0}$ &  \underline{94.6}$_{1.1}$ &  \underline{95.2}$_{0.6}$ &  \newbftab{92.5}$_{0.2}$ &  \underline{63.6}$_{1.2}$ &  \newbftab{95.2}$_{0.3}$ &  \newbftab{60.8}$_{0.4}$ &  \underline{60.1}$_{0.5}$ &  74.6$_{0.9}$ \\
  mpnet (CA-IH) &  512 &         67.7$_{0.2}$ &  \newbftab{90.4}$_{0.3}$ &  \underline{32.8}$_{0.6}$ &          68.0$_{1.6}$ &         94.9$_{0.6}$ &  \underline{94.4}$_{1.5}$ &         90.1$_{1.1}$ &  \underline{63.7}$_{1.4}$ &         94.6$_{0.2}$ &         59.5$_{0.7}$ &  \underline{59.7}$_{0.9}$ &  74.2$_{0.9}$ \\
     mpnet (SN) &  512 &  \newbftab{68.9}$_{0.2}$ &  \underline{90.3}$_{0.3}$ &  \underline{33.2}$_{0.3}$ &   \newbftab{74.3}$_{0.9}$ &  \newbftab{96.1}$_{0.3}$ &  \newbftab{95.3}$_{0.6}$ &         84.0$_{0.3}$ &  \newbftab{64.6}$_{0.7}$ &         90.0$_{0.3}$ &         55.3$_{0.3}$ &  \newbftab{60.4}$_{0.5}$ &  73.9$_{0.5}$ \\
  mpnet (SN-IH) &  512 &  \newbftab{68.9}$_{0.2}$ &  \underline{90.2}$_{0.2}$ &  \newbftab{33.5}$_{0.5}$ &  \underline{62.8}$_{19.6}$ &  \underline{95.9}$_{0.4}$ &  \underline{95.0}$_{0.6}$ &         83.7$_{0.3}$ &  \underline{64.1}$_{0.8}$ &         90.1$_{0.2}$ &         55.1$_{0.3}$ &  \underline{60.3}$_{0.6}$ &  72.7$_{5.9}$ \\
\bottomrule
\end{tabular}
}
\caption{English results for models based on \mpnet and trained on \snli and \mnli, comparing Siamese architecture (SN) and cross attention (CA) and also models with a identity hypothesis (IH). Results are grouped by the number of training examples (n). \underline{Underlined} results are significant. \textbf{Bold} font indicates maxima.}
\label{tab:results_ca}
\end{table*}

\begin{table*}[t]
\centering
\resizebox{\textwidth}{!}{%
\begin{tabular}{lrllll@{\hskip 0.4in}lll@{\hskip 0.4in}lll@{\hskip 0.4in}l}
\toprule
language && \multicolumn{4}{l}{German} & \multicolumn{3}{l}{English} & \multicolumn{3}{l}{Spanish} \\
                 name &    n &                \gnad &       \amazonreviews &             \deisear &              \sbtenk &       \amazonreviews &             \semeval &             \unified &       \amazonreviews &              \headqa &               \sab s &          Mean \\
\midrule
               random &    0 &                 11.1 &                 20.0 &                 14.3 &                 33.3 &                 20.0 &                 33.3 &                 10.0 &                 20.0 &                 16.7 &                 33.3 &          21.2 \\
             FastText &    0 &         17.3$_{1.0}$ &         15.4$_{0.5}$ &         22.2$_{2.1}$ &         31.5$_{1.5}$ &         18.6$_{0.5}$ &  \newbftab{43.8}$_{0.4}$ &         11.8$_{0.3}$ &         19.7$_{0.5}$ &  \newbftab{45.0}$_{0.9}$ &  \underline{35.0}$_{2.2}$ &  26.0$_{1.2}$ \\
     xlm-roberta (CA) &    0 &         28.5$_{1.3}$ &         24.4$_{0.6}$ &         21.1$_{1.8}$ &         34.1$_{1.4}$ &         23.8$_{0.5}$ &         33.1$_{0.2}$ &  \newbftab{16.5}$_{0.3}$ &         24.1$_{0.5}$ &         36.7$_{0.9}$ &         29.5$_{2.2}$ &  27.2$_{1.2}$ \\
 xlm-roberta (CA-IH) &    0 &         29.4$_{1.3}$ &         26.1$_{0.6}$ &         18.3$_{1.5}$ &         31.8$_{0.9}$ &         29.2$_{0.6}$ &         34.6$_{0.2}$ &         15.7$_{0.4}$ &         25.0$_{0.5}$ &         37.8$_{0.9}$ &         24.3$_{1.5}$ &  27.2$_{1.0}$ \\
     xlm-roberta (SN) &    0 &  \newbftab{41.5}$_{1.2}$ &  \newbftab{31.1}$_{0.7}$ &         22.1$_{1.9}$ &  \newbftab{38.4}$_{1.2}$ &  \newbftab{37.0}$_{0.6}$ &         43.1$_{0.3}$ &         15.3$_{0.3}$ &         28.0$_{0.6}$ &         35.4$_{0.9}$ &         32.0$_{2.3}$ &  32.4$_{1.2}$ \\
 xlm-roberta (SN-IH) &    0 &         38.9$_{1.2}$ &         29.5$_{0.5}$ &  \underline{23.0}$_{2.4}$ &         35.7$_{1.4}$ &         31.0$_{0.6}$ &         38.7$_{0.3}$ &         13.7$_{0.2}$ &  \newbftab{32.9}$_{0.6}$ &         38.8$_{0.8}$ &  \newbftab{35.6}$_{2.3}$ &  31.8$_{1.3}$ \\
\midrule
             Char-SVM &    8 &         56.1$_{2.8}$ &         30.5$_{2.2}$ &         29.4$_{1.6}$ &         45.4$_{2.5}$ &         30.0$_{1.6}$ &         33.6$_{1.1}$ &         12.2$_{1.1}$ &         30.8$_{1.2}$ &         36.3$_{2.6}$ &  \underline{50.6}$_{5.3}$ &  35.5$_{2.5}$ \\
     xlm-roberta (CA) &    8 &  \underline{61.6}$_{2.4}$ &  \underline{43.3}$_{1.3}$ &  \newbftab{39.5}$_{5.1}$ &  \underline{53.6}$_{2.2}$ &  \underline{41.2}$_{2.2}$ &  \underline{55.0}$_{3.4}$ &  \underline{18.3}$_{1.4}$ &  \underline{41.1}$_{1.3}$ &  \underline{49.5}$_{2.7}$ &  \underline{53.9}$_{3.5}$ &  45.7$_{2.8}$ \\
 xlm-roberta (CA-IH) &    8 &  \underline{60.2}$_{2.3}$ &  \newbftab{43.9}$_{1.2}$ &  \underline{36.4}$_{1.8}$ &  \newbftab{56.5}$_{1.5}$ &  \underline{43.5}$_{2.0}$ &  \newbftab{55.8}$_{2.9}$ &  \newbftab{18.8}$_{2.2}$ &  \newbftab{42.7}$_{1.7}$ &  \underline{47.6}$_{2.6}$ &  \newbftab{56.5}$_{3.2}$ &  46.2$_{2.2}$ \\
     xlm-roberta (SN) &    8 &  \newbftab{62.8}$_{0.6}$ &         40.0$_{0.9}$ &  \underline{35.2}$_{3.0}$ &         52.6$_{0.6}$ &  \newbftab{43.6}$_{0.6}$ &  \underline{55.6}$_{2.3}$ &  \underline{18.5}$_{0.9}$ &  \underline{40.8}$_{2.8}$ &  \newbftab{50.3}$_{1.2}$ &  \underline{54.6}$_{3.6}$ &  45.4$_{2.0}$ \\
 xlm-roberta (SN-IH) &    8 &         59.2$_{1.5}$ &         41.5$_{1.3}$ &  \underline{33.8}$_{2.4}$ &         53.4$_{1.3}$ &  \underline{43.2}$_{0.9}$ &  \underline{51.8}$_{3.6}$ &  \underline{17.2}$_{0.8}$ &  \underline{41.4}$_{1.4}$ &  \underline{50.2}$_{1.2}$ &  \underline{52.6}$_{4.5}$ &  44.4$_{2.2}$ \\
\midrule
             Char-SVM &   64 &  \underline{77.3}$_{0.8}$ &         41.4$_{0.8}$ &         48.1$_{2.9}$ &         51.5$_{0.7}$ &         43.5$_{0.8}$ &         39.0$_{0.8}$ &         17.3$_{0.4}$ &         40.4$_{1.0}$ &         52.3$_{0.8}$ &         54.7$_{0.9}$ &  46.6$_{1.2}$ \\
     xlm-roberta (CA) &   64 &  \newbftab{78.4}$_{1.1}$ &  \newbftab{51.0}$_{1.6}$ &         56.8$_{1.6}$ &  \newbftab{65.6}$_{0.8}$ &  \underline{51.2}$_{1.5}$ &  \newbftab{61.9}$_{1.1}$ &  \underline{24.3}$_{1.7}$ &  \newbftab{49.5}$_{0.7}$ &  \underline{55.0}$_{0.7}$ &  \underline{61.4}$_{2.0}$ &  55.5$_{1.3}$ \\
 xlm-roberta (CA-IH) &   64 &  \underline{78.3}$_{1.4}$ &  \underline{50.8}$_{1.5}$ &  \underline{57.2}$_{2.0}$ &  \underline{64.3}$_{1.4}$ &  \newbftab{51.3}$_{1.3}$ &  \underline{61.6}$_{0.5}$ &  \newbftab{24.6}$_{1.0}$ &  \underline{48.4}$_{1.6}$ &  \underline{56.0}$_{1.6}$ &  \underline{60.7}$_{2.4}$ &  55.3$_{1.6}$ \\
     xlm-roberta (SN) &   64 &  \underline{77.4}$_{0.6}$ &  \underline{49.6}$_{0.8}$ &  \newbftab{59.3}$_{1.1}$ &         58.8$_{2.3}$ &  \underline{49.7}$_{1.6}$ &         58.3$_{2.1}$ &  \underline{23.6}$_{0.7}$ &         47.3$_{0.4}$ &  \underline{56.0}$_{0.8}$ &  \newbftab{61.8}$_{2.7}$ &  54.2$_{1.5}$ \\
 xlm-roberta (SN-IH) &   64 &  \underline{77.0}$_{0.9}$ &  \underline{49.8}$_{0.9}$ &         56.8$_{0.6}$ &         60.3$_{1.3}$ &  \underline{49.8}$_{1.4}$ &         57.5$_{1.8}$ &         22.8$_{0.8}$ &         46.8$_{0.3}$ &  \newbftab{56.3}$_{1.1}$ &  \underline{59.5}$_{2.7}$ &  53.6$_{1.3}$ \\
\midrule
             Char-SVM &  512 &  \underline{85.0}$_{0.3}$ &         48.2$_{0.5}$ &         48.1$_{2.9}$ &         59.0$_{0.4}$ &         50.4$_{0.4}$ &         46.0$_{0.5}$ &         23.0$_{0.4}$ &         46.4$_{0.9}$ &         64.7$_{0.4}$ &         63.8$_{1.3}$ &  53.5$_{1.1}$ \\
     xlm-roberta (CA) &  512 &  \underline{84.7}$_{0.7}$ &  \underline{56.3}$_{0.3}$ &  \underline{56.5}$_{1.9}$ &  \newbftab{68.5}$_{1.6}$ &  \newbftab{58.6}$_{0.8}$ &  \newbftab{62.7}$_{0.6}$ &  \underline{29.2}$_{0.7}$ &  \newbftab{53.0}$_{0.4}$ &  \underline{65.9}$_{1.0}$ &  \underline{67.9}$_{0.6}$ &  60.3$_{1.0}$ \\
 xlm-roberta (CA-IH) &  512 &  \newbftab{85.8}$_{1.3}$ &  \newbftab{56.8}$_{0.7}$ &         56.3$_{1.8}$ &  \underline{67.9}$_{1.4}$ &  \underline{58.5}$_{0.5}$ &  \underline{62.5}$_{1.3}$ &  \underline{28.9}$_{1.0}$ &  \underline{52.3}$_{1.4}$ &         65.9$_{0.5}$ &  \newbftab{68.9}$_{1.2}$ &  60.4$_{1.2}$ \\
     xlm-roberta (SN) &  512 &  \underline{85.0}$_{0.6}$ &         55.7$_{0.4}$ &  \newbftab{59.5}$_{1.8}$ &  \underline{67.9}$_{0.5}$ &  \newbftab{58.6}$_{0.4}$ &  \underline{62.3}$_{0.8}$ &  \newbftab{29.5}$_{0.4}$ &  \underline{52.5}$_{0.8}$ &  \newbftab{66.9}$_{0.1}$ &         65.6$_{1.1}$ &  60.3$_{0.8}$ \\
 xlm-roberta (SN-IH) &  512 &  \underline{84.9}$_{0.5}$ &  \underline{56.1}$_{0.5}$ &  \underline{57.6}$_{0.9}$ &  \underline{67.8}$_{1.5}$ &  \underline{58.3}$_{0.2}$ &         61.3$_{0.9}$ &  \underline{29.1}$_{0.6}$ &  \underline{52.4}$_{0.6}$ &  \underline{66.8}$_{0.9}$ &  \underline{68.3}$_{1.2}$ &  60.3$_{0.9}$ \\
\bottomrule
\end{tabular}
}
\caption{Multi-lingual results for models based on roberta-xlm for cross attention (CA) and Siamese networks (SN). $n$ denotes the number of training examples. \underline{Underlined} results are significant. \textbf{Bold} font indicates maxima.}
\label{tab:results_ml}
\end{table*}

\begin{table*}[t]
\centering
\resizebox{\textwidth}{!}{%
\begin{tabular}{lrllllllllllll}
\toprule
             name &    n &               \yahoo &              \agnews &             \unified &                \cola &                \subj &                \trec &                \imdb &             \semeval &             \yelp pol &            \yelp full &              \amazonreviews &          Mean \\
\midrule
      mpnet &    0 &         55.0$_{0.2}$ &         65.6$_{0.4}$ &         20.5$_{0.3}$ &  \underline{47.6}$_{1.4}$ &         62.8$_{0.9}$ &         43.0$_{2.1}$ &         79.5$_{0.2}$ &  \newbftab{48.9}$_{0.3}$ &         79.9$_{0.2}$ &  \underline{32.1}$_{0.2}$ &         37.0$_{0.7}$ &  52.0$_{0.9}$ \\
      mpnet (LR) &    0 &  \newbftab{59.1}$_{0.2}$ &  \newbftab{73.8}$_{0.5}$ &  \newbftab{20.9}$_{0.3}$ &  \newbftab{47.7}$_{1.5}$ &  \newbftab{68.7}$_{0.8}$ &  \newbftab{48.2}$_{2.2}$ &  \newbftab{80.0}$_{0.2}$ &         46.3$_{0.3}$ &  \newbftab{80.5}$_{0.2}$ &         28.6$_{0.2}$ &  \newbftab{39.8}$_{0.6}$ &  54.0$_{0.9}$ \\
\midrule
  mpnet (BitFit) &    8 &  \underline{62.6}$_{0.7}$ &         80.1$_{1.5}$ &  \newbftab{27.0}$_{1.2}$ &  \underline{49.0}$_{0.9}$ &  \underline{79.6}$_{3.0}$ &         57.9$_{2.0}$ &  \newbftab{83.9}$_{0.9}$ &  \underline{54.6}$_{2.8}$ &  \underline{90.3}$_{1.9}$ &  \underline{50.1}$_{1.4}$ &  \underline{46.1}$_{1.2}$ &  61.9$_{1.7}$ \\
      mpnet (FT) &    8 &  \underline{63.5}$_{0.8}$ &  \underline{83.3}$_{1.9}$ &  \newbftab{27.0}$_{0.8}$ &  \newbftab{49.7}$_{0.9}$ &  \underline{83.1}$_{4.8}$ &  \newbftab{70.8}$_{7.1}$ &  \underline{82.6}$_{2.3}$ &  \underline{54.8}$_{3.3}$ &  \newbftab{90.6}$_{1.1}$ &  \underline{50.5}$_{1.6}$ &  \newbftab{46.8}$_{1.6}$ &  63.9$_{3.0}$ \\
   mpnet (FT+LR) &    8 &  \newbftab{63.9}$_{1.0}$ &  \newbftab{83.6}$_{1.8}$ &  \underline{26.3}$_{0.8}$ &  \underline{49.1}$_{1.1}$ &  \underline{84.5}$_{3.4}$ &  \underline{68.9}$_{7.3}$ &  \underline{83.6}$_{2.5}$ &  \newbftab{56.9}$_{1.5}$ &  \underline{90.5}$_{1.2}$ &  \newbftab{51.1}$_{1.2}$ &  \underline{46.7}$_{1.9}$ &  64.1$_{2.8}$ \\
      mpnet (LR) &    8 &         59.7$_{0.3}$ &         76.0$_{0.6}$ &         22.4$_{0.4}$ &         47.8$_{0.5}$ &         71.3$_{1.4}$ &         48.4$_{2.7}$ &         80.4$_{0.3}$ &         50.9$_{2.0}$ &         81.7$_{1.5}$ &         33.6$_{3.8}$ &         41.2$_{1.5}$ &  55.8$_{1.7}$ \\
      mpnet (LT) &    8 &         59.4$_{0.9}$ &         78.7$_{0.9}$ &         23.2$_{0.4}$ &  \underline{48.7}$_{1.4}$ &  \underline{81.9}$_{3.4}$ &         52.5$_{4.4}$ &         77.7$_{0.5}$ &         45.2$_{2.0}$ &         85.1$_{2.2}$ &         41.5$_{1.1}$ &         41.9$_{2.9}$ &  57.8$_{2.2}$ \\
 mpnet (LT-DIST) &    8 &  \underline{62.9}$_{0.7}$ &  \underline{83.0}$_{1.9}$ &  \underline{26.6}$_{0.9}$ &  \underline{47.7}$_{3.0}$ &  \newbftab{84.6}$_{3.4}$ &  \underline{67.8}$_{6.4}$ &  \underline{83.7}$_{0.6}$ &  \underline{54.9}$_{2.2}$ &  \underline{89.9}$_{1.4}$ &         49.2$_{1.0}$ &  \underline{45.6}$_{2.1}$ &  63.3$_{2.7}$ \\
\midrule
  mpnet (BitFit) &   64 &  \newbftab{67.6}$_{0.6}$ &  \underline{86.9}$_{0.9}$ &  \newbftab{30.3}$_{0.9}$ &         51.3$_{0.9}$ &  \underline{93.7}$_{0.9}$ &         82.1$_{2.9}$ &  \underline{85.7}$_{1.0}$ &  \underline{60.8}$_{1.4}$ &  \newbftab{92.1}$_{0.5}$ &  \newbftab{54.9}$_{0.7}$ &  \underline{51.8}$_{1.2}$ &  68.8$_{1.3}$ \\
      mpnet (FT) &   64 &  \underline{67.3}$_{0.5}$ &  \underline{87.3}$_{1.2}$ &  \underline{29.5}$_{0.4}$ &  \underline{55.4}$_{1.2}$ &  \newbftab{93.8}$_{0.5}$ &  \newbftab{88.5}$_{2.6}$ &  \underline{86.1}$_{1.2}$ &  \newbftab{61.4}$_{3.0}$ &  \underline{91.8}$_{0.3}$ &  \underline{54.5}$_{0.4}$ &  \underline{53.6}$_{1.6}$ &  69.9$_{1.5}$ \\
   mpnet (FT+LR) &   64 &  \underline{67.5}$_{0.4}$ &  \newbftab{87.6}$_{0.8}$ &  \underline{29.4}$_{0.3}$ &  \newbftab{55.5}$_{0.9}$ &  \underline{93.7}$_{0.5}$ &  \underline{86.5}$_{3.4}$ &  \newbftab{86.2}$_{0.4}$ &  \underline{60.4}$_{2.1}$ &  \underline{91.4}$_{0.6}$ &  \underline{54.6}$_{0.8}$ &  \newbftab{54.1}$_{1.6}$ &  69.7$_{1.4}$ \\
      mpnet (LR) &   64 &         59.9$_{0.1}$ &         76.6$_{0.3}$ &         22.7$_{0.2}$ &         47.8$_{0.5}$ &         71.6$_{0.5}$ &         51.1$_{1.0}$ &         80.4$_{0.1}$ &         52.0$_{0.7}$ &         82.1$_{0.7}$ &         29.8$_{1.3}$ &         42.0$_{0.5}$ &  56.0$_{0.7}$ \\
      mpnet (LT) &   64 &         64.8$_{0.3}$ &         85.0$_{0.6}$ &         27.1$_{0.6}$ &         49.3$_{1.2}$ &         89.9$_{0.5}$ &         70.8$_{2.8}$ &         81.2$_{1.0}$ &         54.5$_{2.7}$ &         89.0$_{0.6}$ &         50.0$_{0.7}$ &         49.1$_{1.6}$ &  64.6$_{1.4}$ \\
 mpnet (LT-DIST) &   64 &  \underline{67.0}$_{0.5}$ &  \underline{86.9}$_{0.9}$ &         28.8$_{0.4}$ &         52.2$_{1.2}$ &         92.5$_{0.2}$ &  \underline{86.5}$_{1.1}$ &         84.6$_{0.3}$ &  \underline{60.2}$_{2.3}$ &         91.2$_{0.3}$ &         53.7$_{0.7}$ &  \underline{52.7}$_{1.2}$ &  68.7$_{1.0}$ \\
\midrule
  mpnet (BitFit) &  512 &  \newbftab{70.4}$_{0.2}$ &         90.3$_{0.2}$ &  \underline{32.9}$_{0.2}$ &  \underline{72.9}$_{1.3}$ &  \newbftab{96.3}$_{0.2}$ &         92.2$_{0.6}$ &  \newbftab{88.2}$_{0.2}$ &  \newbftab{64.4}$_{0.8}$ &  \newbftab{93.3}$_{0.2}$ &  \newbftab{58.5}$_{0.2}$ &         60.7$_{0.3}$ &  74.5$_{0.5}$ \\
      mpnet (FT) &  512 &         69.3$_{0.2}$ &  \underline{90.7}$_{0.3}$ &  \newbftab{33.0}$_{0.4}$ &  \newbftab{74.5}$_{1.2}$ &  \underline{96.0}$_{0.2}$ &  \newbftab{95.4}$_{1.3}$ &  \underline{87.7}$_{0.4}$ &  \underline{64.1}$_{0.8}$ &  \underline{93.2}$_{0.3}$ &  \newbftab{58.5}$_{0.2}$ &  \underline{60.8}$_{0.7}$ &  74.8$_{0.7}$ \\
   mpnet (FT+LR) &  512 &         69.5$_{0.2}$ &  \newbftab{90.8}$_{0.3}$ &  \underline{32.6}$_{0.5}$ &  \underline{74.2}$_{0.9}$ &  \newbftab{96.3}$_{0.3}$ &  \underline{95.0}$_{0.9}$ &  \underline{88.0}$_{0.6}$ &  \underline{63.3}$_{0.7}$ &  \newbftab{93.3}$_{0.2}$ &  \underline{58.4}$_{0.2}$ &  \newbftab{61.3}$_{0.3}$ &  74.8$_{0.5}$ \\
      mpnet (LR) &  512 &         60.1$_{0.1}$ &         76.7$_{0.2}$ &         22.6$_{0.1}$ &         47.8$_{0.3}$ &         72.0$_{0.2}$ &         51.4$_{0.3}$ &         80.3$_{0.0}$ &         52.6$_{0.2}$ &         81.5$_{0.2}$ &         29.7$_{0.3}$ &         42.7$_{0.2}$ &  56.1$_{0.2}$ \\
      mpnet (LT) &  512 &         68.0$_{0.2}$ &         88.0$_{0.3}$ &         29.1$_{0.4}$ &         55.2$_{1.1}$ &         92.6$_{0.5}$ &         86.2$_{0.2}$ &         84.3$_{0.3}$ &         59.8$_{0.7}$ &         91.0$_{0.2}$ &         53.7$_{0.3}$ &         54.9$_{0.5}$ &  69.3$_{0.5}$ \\
 mpnet (LT-DIST) &  512 &         68.7$_{0.2}$ &         88.9$_{0.2}$ &         30.8$_{0.3}$ &         58.6$_{1.1}$ &         93.7$_{0.2}$ &         89.4$_{0.5}$ &         85.5$_{0.2}$ &         61.3$_{0.5}$ &         91.7$_{0.1}$ &         55.8$_{0.2}$ &         57.0$_{0.6}$ &  71.0$_{0.5}$ \\
\bottomrule
\end{tabular}
}
\caption{English results for Siamese models based on \mpnet and trained on NLI and paraphrasing datasets. Comparing fine-tuning (FT), label tuning (LT), label tuning with distillation (LT-DIST), and label refinement (LR). Results are grouped by the number of training examples (n). \underline{Underlined} results are significant. \textbf{Bold} font indicates maxima.}
\label{tab:results_sft}
\end{table*}

\subsection{Models}

\paragraph{Random} The theoretical performance of a random model that uniformly samples labels from the label set.

\paragraph{Word embeddings}
For the English experiments, we use Word2Vec \citep{mikolov2013} embeddings\footnote{\url{https://code.google.com/archive/p/word2vec}}.
For the multi-lingual experiments, we use FastText \citep{grave2018learning}.
In all cases we preprocess using the NLTK tokenizer \citep{nltk} and stop-words list and by filtering non-alphabetic tokens.
% We compute the sentence embeddings by averaging the individual token embeddings.
Sentence embeddings are computed by averaging the token embeddings.
% We also experimented with scaling the embeddings with their inverse unigram frequency but did not find this to improve performance.

\paragraph{Char-SVM}
For the few-shot experiments we implemented a Support Vector Machines (SVM) \cite{hearst1998support}
based on character n-grams.
The model was implemented using the text vectorizer of scikit-learn \citep{scikit-learn} and uses bigrams to fivegrams.

\paragraph{Cross Attention}
For our experiments we use pre-trained models from HuggingFace \citep{wolf-etal-2020-transformers}.
As the cross attention baseline, 
we  trained a version of MPNET \citep{NEURIPS2020_c3a690be} on
Multi-Genre (MNLI, \citet{N18-1101}) and Stanford NLI (SNLI, \citet{bowman-etal-2015-large}) using the parameters and code of \citet{nie-etal-2020-adversarial}.
This model has approx. 110M parameters.
For the multilingual experiments, we trained -- the cross-lingual language model -- XLM roberta-base \citep{DBLP:journals/corr/abs-1907-11692} on SNLI, MNLI, adversarial NLI (ANLI, \citet{nie-etal-2020-adversarial}) and cross-lingual NLI (XNLI, \citet{conneau2018xnli}), using the same code and parameters as above. The model has approx. 280M parameters.
We give more details on the NLI datasets in Appendix \ref{sec:nli_train}.

\paragraph{Siamese Network}
We also use models based on MPNET for the experiments with the Siamese Networks.
\textit{paraphrase-mpnet-base-v2}\footnote{\url{https://tinyurl.com/pp-mpnet}} is a sentence transformer model
\citep{reimers-2019-sentence-bert} trained on a variety of paraphrasing datasets as well as SNLI and MNLI using a batch softmax loss \citep{HendersonASSLGK17}.
% The sentence embedding is computed by averaging the contextualized token embeddings (mean pooling).
\textit{nli-mpnet-base-v2}\footnote{\url{https://tinyurl.com/nli-mpnet}} is identical to the previous model but trained exclusively on MNLI and SNLI and thus comparable to the cross attention model.
For the multilingual experiments, we trained a model using the code of the sentence transformers with the same batch softmax objective used for fine-tuning the few-shot models
and on the same data we used for training the cross attention model.

\paragraph{Roberta-NatCat}
For comparison with the related work, we also trained a model based on \roberta{} \citep{DBLP:journals/corr/abs-1907-11692} and fine-tuned on the NatCat dataset as discussed in \citet{chu2020} using the code\footnote{\url{https://github.com/ZeweiChu/ULR}} and parameters of the authors.

\mysubsection{Datasets}
We use a number of English text classification datasets used in the zero-shot and the few-shot literature \citep{yin-etal-2019-benchmarking,gao2020,Wang2021EntailmentAF}.
In addition, we use several German and Spanish datasets for the multilingual experiments.
Table \ref{tab:dataset_stats} provides more details.

These datasets are of a number of common text classification tasks such as topic classification,
sentiment and emotion detection, and review rating.
However, we also included some less well-known tasks such as acceptability, whether an English sentence is deemed acceptable by a native speaker,
and subjectivity, whether a statement is subjective or objective.
As some datasets do not have a standard split we split them randomly using a 9/1 ratio.
% The split is done using a hash function based on the text.

\mysubsection{Hypotheses}
We use the same hypotheses for the cross attention model and for the Siamese network.
For \yahoo and \unified we use the hypotheses from \citet{yin-etal-2019-benchmarking}.
For \subj, \cola, \trec, \yelp, \agnews and \imdb we use the same hypotheses as \citet{Wang2021EntailmentAF}.
For the remaining datasets we designed our own hypotheses.
These were written in an attempt to mirror what has been done for other datasets and they have not been tuned in any way.
Appendix \ref{sec:entail_patterns} shows the patterns used.
We also explored using an identity hypothesis, that is the raw label names as the label representation and found this to give similar results.

\mysubsection{Fine-Tuning}
Inspired by \citet{Wang2021EntailmentAF}, we investigate fine-tuning the models with 8, 64 and 512 examples per label.
For fine-tuning the cross attention models we follow the literature \citep{Wang2021EntailmentAF} and create examples of every possible combination of input text and label.
The example corresponding to the correct label is labeled as entailed while all other examples are labeled as refuted.
We then fine-tune the model using stochastic gradient descent and a cross-entropy loss.
We use a learning rate of 1e-5, a batch size of 8 and run the training for 10 epochs.
As discussed in the methodology Section~\ref{sec:finetuning}, for the Siamese Networks every batch contains exactly one example of every label and therefore the batch size equals the number of labels of the task.
We use a learning rate of 2e-5 and of 2e-4 for the \bitfit experiments.
Appendix \ref{sec:hparams} contains additional information on the hyper-parameters used.

% To compensate for the noise introduced by the small training set, 
We use macro F1-score as the evaluation metric.
We run all experiments with 5 different training sets and report the mean and standard deviation.
For the zero-shot experiments, we estimate the standard deviation using bootstrapping \citep{koehn-2004-statistical}.
In all cases, we use Welch's t-test\footnote{\url{https://en.wikipedia.org/wiki/Welch\%27s_t-test}} with a p-value of 0.05 to establish significance (following \citet{logan2021}).
For the experiments with label refinement \citep{chu2020} and distillation, we use up to 10,000 unlabeled examples from the training set.

\section{Results}

Here we present the results of our experiments.
The two main questions we want to answer are whether Siamese Networks (SN) give comparable results as Cross Attention models (CA) and how well Label Tuning (LT) compares to regular fine-tuning.

\mysubsection{Siamese Network and Cross Attention}

Table \ref{tab:results_ca} shows results comparing SN with CA and various baselines.
As discussed above, SN and CA models are based on the MPNET architecture and trained on SNLI and MNLI.

For the zero-shot setup ($n{=}0$) we see that all models out-perform the random baseline on average.
The word embedding baselines and \robertanatcat perform significantly worse than random on several of the datasets.
In contrast the SN and CA models only perform worse than random on \cola.
The SN outperforms the CA on average, but the results for the individual datasets are mixed. 
The SN is significantly better for 4, significantly worse for 4 and on par for the remaining 3 datasets.
Regarding the use of a hypothesis pattern from the literature or just an identity hypothesis (IH), we find that,
while there are significant differences on individual datasets, the IH setup shows higher but still comparable (within 1 point) average performance.

For the few-shots setup ($n{=}\{8,64,512\}$), we find that all models out-perform 
a Char-SVM trained with the same number of instances by a large margin.
Comparing SN and CA, we see that CA outperforms the SN on average but with a difference with-in the confidence interval.
For $n{=}8$ and $n{=}64$, CA significantly outperforms SN on 3 datasets and performs comparably on the remaining 8.
For $n{=}512$, we see an even more mixed picture. CA is on par with SN on 6 datasets, outperforms it on 3 and is out-performed on 2.
We can conclude that for the English datasets, SN is more accurate for zero-shot while CA is more accurate for few-shot.
The average difference is small in both setups and we do not see a significant difference for most datasets.

Table \ref{tab:results_ml} shows the multi-lingual experiments.
The \roberta{} XLM models were pre-trained on data from more than 100 languages and fine-tuned on an NLI data of 15 languages.
The cross-lingual data and the fact that there is only 7500 examples for the languages other than English, explains why quality is lower than for the English-only experiments. 
For the zero-shot scenario, all models out-perform the random baseline on average, but with a smaller margin than for the English-only models.
The FastText baseline performs comparable to CA on average (26.0 vs 27.2), while SN is ahead by a large margin (27.2 vs 32.4).
The differences between models with hypotheses and identity hypothesis (IH) are smaller than for the English experiments.

Looking at the few-shot scenarios, we see that both models out-perform the Char-SVM by a large margin.
In general, the results are closer than for the English experiments, as well as in the number of datasets
with significant differences (only 2-4 of  datasets). 
Similarly to English, we can conclude that at multilingual level, SN is more accurate in the zero-shot scenario whereas CA performs better in the few-shot one.
However, for few-shot we see only small average differences (less than 1 point except for $n{=}64$).

\mysubsection{Label Tuning}

Table \ref{tab:results_sft} shows a comparison of different fine-tuning approaches on the English datasets.
Appendix \ref{sec:results_sft_ml} contains the multi-lingual results and gives a similar picture.
We first compare Label Refinement (LR) as discussed in \citet{chu2020} (see Section~\ref{lr:entailemnt}).
Recall that this approach makes use of unlabeled data.
We find that in the zero-shot scenario LR gives an average improvement of more than 2 points and significantly out-performing the baseline (mpnet) for 7 of the 11 datasets.
When combining LR with labeled data as discussed in \citet{chu2020} we find this to only give modest improvements over the zero-shot model (e.g., 54.0 (zero-shot) vs 55.8 ($n{=}8$)).
Note that we apply LR to the untuned model, while \citet{chu2020} proposed to apply it to a tuned model.
However, we find that to only give small improvements over an already tuned model (mpnet (FT) vs. mpnet (FT+LR)). 
Also, in this work we are interested in approaches that do not change the initial model so that it can be shared between tasks to improve scalability. 
Label Tuning (LT) improves results as $n$ grows and out-performs LR and the Char-SVM baseline from Table \ref{tab:results_ca}. 

Comparing regular Fine-Tuning (FT) and BitFit, we find them to perform quite similarly both on average and on individual datasets, with only few exceptions, such as the performance difference on \trec for the $n{=}8$ setup.
In comparison with FT and \bitfit, LT is significantly out-performed on most datasets. The average difference in performance is around 5 points, which is comparable to using 8 times less training data.

Using the knowledge distillation approach discussed before (LT-DIST), we find that for 8 and 64 examples, most of the difference in performance can be recovered while still keeping the high scalability. For $n{=}8$, we only find a significant difference to mpnet (FT) for \yelp full. Recall that the distillation is performed on up to 10,000 unlabeled examples from the training set.

\begin{table}[t]
\begin{center}
\resizebox{0.8\columnwidth}{!}{%
\begin{tabular}{lrrr}
\toprule
               name &       2-3 &       4-6 &      10 \\
\midrule
 W2V &  192.90 &  195.82 &  208.40 \\
 mpnet-base (CA) &    5.12 &    2.22 &    1.15 \\
 mpnet-base (SN) &   26.08 &   18.30 &   18.85 \\
\bottomrule
\end{tabular}
}%
\end{center}
\caption{Processing speed in \textit{thousand tokens/second}. We show the results grouped by the size of the label set. Calculated on the English test sets.}
\label{tab:speed}
\end{table}

\begin{table}
\resizebox{\columnwidth}{!}{%
\begin{tabular}{lrrrrr}
\toprule
length &  1-22 &  22-44 &  44-86 &  86-160 &  $>$ 160 \\
\midrule
  SN &  39.8 &   44.6 &   42.5 &    34.5 &  36.4 \\
  CA &  36.7 &   41.8 &   44.0 &    35.2 &  40.3 \\
\bottomrule
\end{tabular}
}%
\caption{Average macro F1 score for sets of different token length measured across all test sets for $n{=}0$.}
\label{tab:a_length}
\end{table}

\begin{table}
\resizebox{\columnwidth}{!}{%
\begin{tabular}{lrrrrrr}
\toprule
task &  \multicolumn{2}{c}{emotions} & \multicolumn{2}{c}{reviews} & \multicolumn{2}{c}{sentiment} \\ 
negation &  no &  yes &  no &  yes &  no &  yes \\
\midrule
  SN &            23.0 &             14.3 &           49.0 &            44.4 &             37.3 &              45.1 \\
  CA &            22.4 &             16.8 &           48.2 &            47.0 &             32.2 &              37.4 \\
\bottomrule
\end{tabular}
}
\caption{Average macro F1 score for sets with and without a negation marker present. Measured across all test sets for $n{=}0$.}
\label{tab:a_negation}
\end{table}

\section{Analysis}

We analyze the performance of the Cross Attention (CA) and Siamese Network-based (SN) models. 
Unless otherwise noted, the analysis was run over all datasets and languages.
% First, in terms of computation time. Next, we extend the study of quality performance with an error analysis from different granularity perspectives.
Table \ref{tab:speed}, gives a comparison of the processing speed of different models.
Details on the hardware used is given in Appendix \ref{sec:hardware}.
As expected, the performance of the cross attention model halves when the label size doubles.
The performance of the Siamese network is independent of the number of labels.
This shows that Siamese Networks have a huge advantage at inference time -- especially for tasks with many labels.

Table \ref{tab:a_length} shows the average F1 scores for different token lengths.
To this end the data was grouped in bins of roughly equal size.
SN has an advantage for shorter sequences ($\leq 44$ tokens), while CA performs better for longer texts ($> 160$ tokens).

Table \ref{tab:a_negation} shows an analysis based on whether the text does or does not contain negation markers.
We used an in-house list of 23 phrases for German and Spanish and 126 for English.
For emotion detection and review tasks, both models perform better on the subset without negations.
However, while SN outperforms CA on the data without negations, CA performs better on the data with negations.
The same trend does not hold for the sentiment datasets.
These are based on Twitter and thus contain shorter and simpler sentences.
For the sentiment datasets based on Twitter we also found that both models struggle to predict the neutral class.
CA classifies almost everything neutral tweet as positive or negative.
SN predicts the neutral class regularly but still with a relative high error rate.
Appendix \ref{sec:analisis} contains further analysis showing that label set size, language and task do not have a visible effect on the difference in accuracy of the two models.

\section{Conclusion}

We have shown that Cross Attention (CA) and Siamese Networks (SN) for zero-shot and few-shot text classification give
comparable results across a diverse set of tasks and multiple languages.
The inference cost of SNs is low as label embeddings can be pre-computed and, in contrast to CA, does not scale with the number of labels.
We also showed that tuning only these label embeddings (Label Tuning (LT)) is an interesting alternative to regular Fine-Tuning (FT).
LT gets close to FT performance when combined with knowledge distillation and when the number of training samples is low, i.e., for realistic few-shot learning.
This is relevant for production scenarios, as it allows to share the same model among tasks.
However, it will require 60 times more memory to add a new task: For a 418 MB mpnet-base model, \bitfit affects 470 kB of the parameters. LT applied to a task with 10 labels and using a embedding dimension of 768 requires 7.5 kB.
The main disadvantage of \bitfit, however, is that the weight sharing it requires is much harder to implement, especially in highly optimized environments such as \href{https://developer.nvidia.com/nvidia-triton-inference-server}{NVIDIA Triton}.
Therefore we think that LT is an interesting alternative for fast and scalable few-shot learning.

\section*{Acknowledgements}

We would like to thank Francisco Rangel and the entire Symanto Research Team for early discussions, feedback and suggestions.
We would also like to thank the anonymous Reviewers.
% http://www.itefi.csic.es/en/proyectos/ia-explicable-para-desinformacion-y-deteccion-de-conspiracion-durante-infodemias
The authors gratefully acknowledge the support of the Pro$^2$Haters - Proactive Profiling of Hate Speech Spreaders (CDTi IDI-20210776), XAI-DisInfodemics: eXplainable AI for disinformation and conspiracy detection during infodemics (MICIN PLEC2021-007681), and DETEMP - Early Detection of Depression Detection in Social Media (IVACE IMINOD/2021/72) R\&D grants.

\bibliography{references}

\begin{thebibliography}{64}
\expandafter\ifx\csname natexlab\endcsname\relax\def\natexlab#1{#1}\fi

\bibitem[{Alm et~al.(2005)Alm, Roth, and Sproat}]{alm2005emotions}
Cecilia~Ovesdotter Alm, Dan Roth, and Richard Sproat. 2005.
\newblock Emotions from text: machine learning for text-based emotion
  prediction.
\newblock In \emph{Proceedings of the conference on human language technology
  and empirical methods in natural language processing}, pages 579--586.
  Association for Computational Linguistics.

\bibitem[{Alm and Sproat(2005)}]{alm2005perceptions}
Cecilia~Ovesdotter Alm and Richard Sproat. 2005.
\newblock Perceptions of emotions in expressive storytelling.
\newblock In \emph{Ninth European Conference on Speech Communication and
  Technology}.

\bibitem[{Alm(2008)}]{alm2008affect}
Ebba Cecilia~Ovesdotter Alm. 2008.
\newblock \emph{Affect in* text and speech}.
\newblock University of Illinois at Urbana-Champaign.

\bibitem[{Baldini~Soares et~al.(2019)Baldini~Soares, FitzGerald, Ling, and
  Kwiatkowski}]{baldini-soares-etal-2019-matching}
Livio Baldini~Soares, Nicholas FitzGerald, Jeffrey Ling, and Tom Kwiatkowski.
  2019.
\newblock \href {https://doi.org/10.18653/v1/P19-1279} {Matching the blanks:
  Distributional similarity for relation learning}.
\newblock In \emph{Proceedings of the 57th Annual Meeting of the Association
  for Computational Linguistics}, pages 2895--2905, Florence, Italy.
  Association for Computational Linguistics.

\bibitem[{Ben-Zaken et~al.(2021)Ben-Zaken, Ravfogel, and
  Goldberg}]{BenZaken202}
Elad Ben-Zaken, Shauli Ravfogel, and Yoav Goldberg. 2021.
\newblock Bitfit: Simple parameter-efficient fine-tuning for transformer-based
  masked language-models.
\newblock \emph{ArXiv}, abs/2106.10199.

\bibitem[{Bird et~al.(2009)Bird, Klein, and Loper}]{nltk}
Steven Bird, Ewan Klein, and Edward Loper. 2009.
\newblock \emph{Natural Language Processing with Python}, 1st edition.
\newblock O'Reilly Media, Inc.

\bibitem[{Block(2019)}]{Gnad}
Timo Block. 2019.
\newblock Ten thousand german news articles dataset.
\newblock \url{https://tblock.github.io/10kGNAD/}.
\newblock Accessed: 2021-08-25.

\bibitem[{Bostan and Klinger(2018)}]{Bostan2018}
Laura Ana~Maria Bostan and Roman Klinger. 2018.
\newblock \href {http://aclweb.org/anthology/C18-1179} {An analysis of
  annotated corpora for emotion classification in text}.
\newblock In \emph{Proceedings of the 27th International Conference on
  Computational Linguistics}, pages 2104--2119. Association for Computational
  Linguistics.

\bibitem[{Bowman et~al.(2015)Bowman, Angeli, Potts, and
  Manning}]{bowman-etal-2015-large}
Samuel~R. Bowman, Gabor Angeli, Christopher Potts, and Christopher~D. Manning.
  2015.
\newblock \href {https://doi.org/10.18653/v1/D15-1075} {A large annotated
  corpus for learning natural language inference}.
\newblock In \emph{Proceedings of the 2015 Conference on Empirical Methods in
  Natural Language Processing}, pages 632--642, Lisbon, Portugal. Association
  for Computational Linguistics.

\bibitem[{Bromley et~al.(1993)Bromley, Guyon, LeCun, S\"{a}ckinger, and
  Shah}]{10.5555/2987189.2987282}
Jane Bromley, Isabelle Guyon, Yann LeCun, Eduard S\"{a}ckinger, and Roopak
  Shah. 1993.
\newblock Signature verification using a "siamese" time delay neural network.
\newblock In \emph{Proceedings of the 6th International Conference on Neural
  Information Processing Systems}, NIPS'93, page 737–744, San Francisco, CA,
  USA. Morgan Kaufmann Publishers Inc.

\bibitem[{Brown et~al.(2020)Brown, Mann, Ryder, Subbiah, Kaplan, Dhariwal,
  Neelakantan, Shyam, Sastry, Askell, Agarwal, Herbert-Voss, Krueger, Henighan,
  Child, Ramesh, Ziegler, Wu, Winter, Hesse, Chen, Sigler, Litwin, Gray, Chess,
  Clark, Berner, McCandlish, Radford, Sutskever, and
  Amodei}]{NEURIPS2020_1457c0d6}
Tom Brown, Benjamin Mann, Nick Ryder, Melanie Subbiah, Jared~D Kaplan, Prafulla
  Dhariwal, Arvind Neelakantan, Pranav Shyam, Girish Sastry, Amanda Askell,
  Sandhini Agarwal, Ariel Herbert-Voss, Gretchen Krueger, Tom Henighan, Rewon
  Child, Aditya Ramesh, Daniel Ziegler, Jeffrey Wu, Clemens Winter, Chris
  Hesse, Mark Chen, Eric Sigler, Mateusz Litwin, Scott Gray, Benjamin Chess,
  Jack Clark, Christopher Berner, Sam McCandlish, Alec Radford, Ilya Sutskever,
  and Dario Amodei. 2020.
\newblock \href
  {https://proceedings.neurips.cc/paper/2020/file/1457c0d6bfcb4967418bfb8ac142f64a-Paper.pdf}
  {Language models are few-shot learners}.
\newblock In \emph{Advances in Neural Information Processing Systems},
  volume~33, pages 1877--1901. Curran Associates, Inc.

\bibitem[{Chang et~al.(2008)Chang, Ratinov, Roth, and
  Srikumar}]{Chang2008ImportanceOS}
Ming-Wei Chang, Lev-Arie Ratinov, D.~Roth, and Vivek Srikumar. 2008.
\newblock Importance of semantic representation: Dataless classification.
\newblock In \emph{AAAI}.

\bibitem[{Chu et~al.(2021)Chu, Stratos, and Gimpel}]{chu2020}
Zewei Chu, Karl Stratos, and Kevin Gimpel. 2021.
\newblock \href {https://doi.org/10.18653/v1/2021.findings-acl.365}
  {Unsupervised label refinement improves dataless text classification}.
\newblock In \emph{Findings of the Association for Computational Linguistics:
  ACL-IJCNLP 2021}, pages 4165--4178, Online. Association for Computational
  Linguistics.

\bibitem[{Cieliebak et~al.(2017)Cieliebak, Deriu, Egger, and
  Uzdilli}]{cieliebak-etal-2017-twitter}
Mark Cieliebak, Jan~Milan Deriu, Dominic Egger, and Fatih Uzdilli. 2017.
\newblock \href {https://doi.org/10.18653/v1/W17-1106} {A {T}witter corpus and
  benchmark resources for {G}erman sentiment analysis}.
\newblock In \emph{Proceedings of the Fifth International Workshop on Natural
  Language Processing for Social Media}, pages 45--51, Valencia, Spain.
  Association for Computational Linguistics.

\bibitem[{Conneau et~al.(2018)Conneau, Rinott, Lample, Williams, Bowman,
  Schwenk, and Stoyanov}]{conneau2018xnli}
Alexis Conneau, Ruty Rinott, Guillaume Lample, Adina Williams, Samuel~R.
  Bowman, Holger Schwenk, and Veselin Stoyanov. 2018.
\newblock Xnli: Evaluating cross-lingual sentence representations.
\newblock In \emph{Proceedings of the 2018 Conference on Empirical Methods in
  Natural Language Processing}. Association for Computational Linguistics.

\bibitem[{Crowdflower(2016)}]{crowdflower2016}
Crowdflower. 2016.
\newblock \href {https://data.world/crowdflower/sentiment-analysis-in-text}
  {The emotion in text, published by crowdflower}.

\bibitem[{Dagan et~al.(2006)Dagan, Glickman, and Magnini}]{10.1007/11736790_9}
Ido Dagan, Oren Glickman, and Bernardo Magnini. 2006.
\newblock The pascal recognising textual entailment challenge.
\newblock In \emph{Machine Learning Challenges. Evaluating Predictive
  Uncertainty, Visual Object Classification, and Recognising Tectual
  Entailment}, pages 177--190, Berlin, Heidelberg. Springer Berlin Heidelberg.

\bibitem[{Devlin et~al.(2019)Devlin, Chang, Lee, and
  Toutanova}]{devlin-etal-2019-bert}
Jacob Devlin, Ming-Wei Chang, Kenton Lee, and Kristina Toutanova. 2019.
\newblock \href {https://doi.org/10.18653/v1/N19-1423} {{BERT}: Pre-training of
  deep bidirectional transformers for language understanding}.
\newblock In \emph{Proceedings of the 2019 Conference of the North {A}merican
  Chapter of the Association for Computational Linguistics: Human Language
  Technologies, Volume 1 (Long and Short Papers)}, pages 4171--4186,
  Minneapolis, Minnesota. Association for Computational Linguistics.

\bibitem[{Gabrilovich and Markovitch(2007)}]{Gabrilovich2007ComputingSR}
Evgeniy Gabrilovich and Shaul Markovitch. 2007.
\newblock Computing semantic relatedness using wikipedia-based explicit
  semantic analysis.
\newblock In \emph{Proceedings of the 20th International Joint Conference on
  Artifical Intelligence}, IJCAI'07, page 1606–1611, San Francisco, CA, USA.
  Morgan Kaufmann Publishers Inc.

\bibitem[{Gao et~al.(2021)Gao, Fisch, and Chen}]{gao2020}
Tianyu Gao, Adam Fisch, and Danqi Chen. 2021.
\newblock \href {https://doi.org/10.18653/v1/2021.acl-long.295} {Making
  pre-trained language models better few-shot learners}.
\newblock In \emph{Proceedings of the 59th Annual Meeting of the Association
  for Computational Linguistics and the 11th International Joint Conference on
  Natural Language Processing (Volume 1: Long Papers)}, pages 3816--3830,
  Online. Association for Computational Linguistics.

\bibitem[{Ghazi et~al.(2015)Ghazi, Inkpen, and Szpakowicz}]{ghazi2016detecting}
Diman Ghazi, Diana Inkpen, and Stan Szpakowicz. 2015.
\newblock Detecting emotion stimuli in emotion-bearing sentences.
\newblock In \emph{CICLing (2)}, pages 152--165.

\bibitem[{Grave et~al.(2018)Grave, Bojanowski, Gupta, Joulin, and
  Mikolov}]{grave2018learning}
Edouard Grave, Piotr Bojanowski, Prakhar Gupta, Armand Joulin, and Tomas
  Mikolov. 2018.
\newblock Learning word vectors for 157 languages.
\newblock In \emph{Proceedings of the International Conference on Language
  Resources and Evaluation (LREC 2018)}.

\bibitem[{Gulli(2005)}]{AgNews}
Antonio Gulli. 2005.
\newblock {AG}'s corpus of news articles.
\newblock
  \url{http://groups.di.unipi.it/~gulli/AG_corpus_of_news_articles.html}.
\newblock Accessed: 2021-07-08.

\bibitem[{Halder et~al.(2020)Halder, Akbik, Krapac, and
  Vollgraf}]{halder-etal-2020-task}
Kishaloy Halder, Alan Akbik, Josip Krapac, and Roland Vollgraf. 2020.
\newblock \href {https://doi.org/10.18653/v1/2020.coling-main.285} {Task-aware
  representation of sentences for generic text classification}.
\newblock In \emph{Proceedings of the 28th International Conference on
  Computational Linguistics}, pages 3202--3213, Barcelona, Spain (Online).
  International Committee on Computational Linguistics.

\bibitem[{Hambardzumyan et~al.(2021)Hambardzumyan, Khachatrian, and
  May}]{hambardzumyan-etal-2021-warp}
Karen Hambardzumyan, Hrant Khachatrian, and Jonathan May. 2021.
\newblock \href {https://doi.org/10.18653/v1/2021.acl-long.381} {{WARP}:
  {W}ord-level {A}dversarial {R}e{P}rogramming}.
\newblock In \emph{Proceedings of the 59th Annual Meeting of the Association
  for Computational Linguistics and the 11th International Joint Conference on
  Natural Language Processing (Volume 1: Long Papers)}, pages 4921--4933,
  Online. Association for Computational Linguistics.

\bibitem[{Hearst et~al.(1998)Hearst, Dumais, Osuna, Platt, and
  Scholkopf}]{hearst1998support}
Marti~A. Hearst, Susan~T Dumais, Edgar Osuna, John Platt, and Bernhard
  Scholkopf. 1998.
\newblock Support vector machines.
\newblock \emph{IEEE Intelligent Systems and their applications}, 13(4):18--28.

\bibitem[{Henderson et~al.(2017)Henderson, Al{-}Rfou, Strope, Sung,
  Luk{\'{a}}cs, Guo, Kumar, Miklos, and Kurzweil}]{HendersonASSLGK17}
Matthew~L. Henderson, Rami Al{-}Rfou, Brian Strope, Yun{-}Hsuan Sung,
  L{\'{a}}szl{\'{o}} Luk{\'{a}}cs, Ruiqi Guo, Sanjiv Kumar, Balint Miklos, and
  Ray Kurzweil. 2017.
\newblock \href {http://arxiv.org/abs/1705.00652} {Efficient natural language
  response suggestion for smart reply}.
\newblock \emph{CoRR}, abs/1705.00652.

\bibitem[{Hinton et~al.(2014)Hinton, Vinyals, and
  Dean}]{Hinton2015DistillingTK}
Geoffrey~E. Hinton, Oriol Vinyals, and J.~Dean. 2014.
\newblock Distilling the knowledge in a neural network.
\newblock In \emph{The {NIPS} 2014 Learning Semantics Workshop}.

\bibitem[{Keung et~al.(2020)Keung, Lu, Szarvas, and
  Smith}]{DBLP:journals/corr/abs-2010-02573}
Phillip Keung, Yichao Lu, Gy{\"o}rgy Szarvas, and Noah~A. Smith. 2020.
\newblock \href {https://doi.org/10.18653/v1/2020.emnlp-main.369} {The
  multilingual {A}mazon reviews corpus}.
\newblock In \emph{Proceedings of the 2020 Conference on Empirical Methods in
  Natural Language Processing (EMNLP)}, pages 4563--4568, Online. Association
  for Computational Linguistics.

\bibitem[{Koehn(2004)}]{koehn-2004-statistical}
Philipp Koehn. 2004.
\newblock \href {https://aclanthology.org/W04-3250} {Statistical significance
  tests for machine translation evaluation}.
\newblock In \emph{Proceedings of the 2004 Conference on Empirical Methods in
  Natural Language Processing}, pages 388--395, Barcelona, Spain. Association
  for Computational Linguistics.

\bibitem[{Larochelle et~al.(2008)Larochelle, Erhan, and
  Bengio}]{larochelle2008}
Hugo Larochelle, Dumitru Erhan, and Yoshua Bengio. 2008.
\newblock Zero-data learning of new tasks.
\newblock In \emph{Proceedings of the 23rd National Conference on Artificial
  Intelligence - Volume 2}, AAAI'08, page 646–651. AAAI Press.

\bibitem[{Le~Scao and Rush(2021)}]{le-scao-rush-2021-many}
Teven Le~Scao and Alexander Rush. 2021.
\newblock \href {https://doi.org/10.18653/v1/2021.naacl-main.208} {How many
  data points is a prompt worth?}
\newblock In \emph{Proceedings of the 2021 Conference of the North American
  Chapter of the Association for Computational Linguistics: Human Language
  Technologies}, pages 2627--2636, Online. Association for Computational
  Linguistics.

\bibitem[{Li and Roth(2002)}]{li-roth-2002-learning}
Xin Li and Dan Roth. 2002.
\newblock \href {https://aclanthology.org/C02-1150} {Learning question
  classifiers}.
\newblock In \emph{{COLING} 2002: The 19th International Conference on
  Computational Linguistics}.

\bibitem[{Li et~al.(2017)Li, Su, Shen, Li, Cao, and Niu}]{li2017dailydialog}
Yanran Li, Hui Su, Xiaoyu Shen, Wenjie Li, Ziqiang Cao, and Shuzi Niu. 2017.
\newblock Dailydialog: A manually labelled multi-turn dialogue dataset.
\newblock \emph{arXiv preprint arXiv:1710.03957}.

\bibitem[{Liu et~al.(2007)Liu, Banea, and Mihalcea}]{Liu17}
V.~Liu, C.~Banea, and R.~Mihalcea. 2007.
\newblock Grounded emotions.
\newblock In \emph{International Conference on Affective Computing and
  Intelligent Interaction (ACII 2017)}, San Antonio, Texas.

\bibitem[{Liu et~al.(2019)Liu, Ott, Goyal, Du, Joshi, Chen, Levy, Lewis,
  Zettlemoyer, and Stoyanov}]{DBLP:journals/corr/abs-1907-11692}
Yinhan Liu, Myle Ott, Naman Goyal, Jingfei Du, Mandar Joshi, Danqi Chen, Omer
  Levy, Mike Lewis, Luke Zettlemoyer, and Veselin Stoyanov. 2019.
\newblock \href {http://arxiv.org/abs/1907.11692} {Roberta: {A} robustly
  optimized {BERT} pretraining approach}.
\newblock \emph{CoRR}, abs/1907.11692.

\bibitem[{Logan~IV et~al.(2021)Logan~IV, Balazevic, Wallace, Petroni, Singh,
  and Riedel}]{logan2021}
Robert~L. Logan~IV, Ivana Balazevic, Eric Wallace, Fabio Petroni, Sameer Singh,
  and Sebastian Riedel. 2021.
\newblock \href
  {https://neurips2021-nlp.github.io/papers/24/CameraReady/NullPrompts___ENLSP_Workshop_Version.pdf}
  {Cutting down on prompts and parameters: Simple few-shot learning with
  language models}.
\newblock In \emph{Advances in Neural Information Processing Systems}.

\bibitem[{Maas et~al.(2011)Maas, Daly, Pham, Huang, Ng, and
  Potts}]{maas-EtAl:2011:ACL-HLT2011}
Andrew~L. Maas, Raymond~E. Daly, Peter~T. Pham, Dan Huang, Andrew~Y. Ng, and
  Christopher Potts. 2011.
\newblock \href {http://www.aclweb.org/anthology/P11-1015} {Learning word
  vectors for sentiment analysis}.
\newblock In \emph{Proceedings of the 49th Annual Meeting of the Association
  for Computational Linguistics: Human Language Technologies}, pages 142--150,
  Portland, Oregon, USA. Association for Computational Linguistics.

\bibitem[{Mikolov et~al.(2013)Mikolov, Chen, Corrado, and Dean}]{mikolov2013}
Tom{\'{a}}s Mikolov, Kai Chen, Greg Corrado, and Jeffrey Dean. 2013.
\newblock \href {http://arxiv.org/abs/1301.3781} {Efficient estimation of word
  representations in vector space}.
\newblock In \emph{1st International Conference on Learning Representations,
  {ICLR} 2013, Scottsdale, Arizona, USA, May 2-4, 2013, Workshop Track
  Proceedings}.

\bibitem[{Mohammad(2012)}]{mohammad-2012-emotional}
Saif Mohammad. 2012.
\newblock \href {https://www.aclweb.org/anthology/S12-1033} {{\#}{E}motional
  {T}weets}.
\newblock In \emph{*{SEM} 2012: The First Joint Conference on Lexical and
  Computational Semantics {--} Volume 1: Proceedings of the main conference and
  the shared task, and Volume 2: Proceedings of the Sixth International
  Workshop on Semantic Evaluation ({S}em{E}val 2012)}, pages 246--255,
  Montr{\'e}al, Canada. Association for Computational Linguistics.

\bibitem[{Mohammad et~al.(2017)Mohammad, Sobhani, and
  Kiritchenko}]{mohammad2017stance}
Saif~M Mohammad, Parinaz Sobhani, and Svetlana Kiritchenko. 2017.
\newblock Stance and sentiment in tweets.
\newblock \emph{ACM Transactions on Internet Technology (TOIT)}, 17(3):1--23.

\bibitem[{Mohammad et~al.(2015)Mohammad, Zhu, Kiritchenko, and
  Martin}]{mohammad2015sentiment}
Saif~M Mohammad, Xiaodan Zhu, Svetlana Kiritchenko, and Joel Martin. 2015.
\newblock Sentiment, emotion, purpose, and style in electoral tweets.
\newblock \emph{Information Processing \& Management}, 51(4):480--499.

\bibitem[{Nakov et~al.(2016)Nakov, Ritter, Rosenthal, Stoyanov, and
  Sebastiani}]{SemEval:2016:task4}
Preslav Nakov, Alan Ritter, Sara Rosenthal, Veselin Stoyanov, and Fabrizio
  Sebastiani. 2016.
\newblock {SemEval}-2016 task 4: Sentiment analysis in {T}witter.
\newblock In \emph{Proceedings of the 10th International Workshop on Semantic
  Evaluation}, SemEval '16, San Diego, California. Association for
  Computational Linguistics.

\bibitem[{Navas-Loro et~al.(2017)Navas-Loro, Rodr{\'i}guez-Doncel,
  Santana-Perez, and S{\'a}nchez}]{10.1007/978-3-319-66429-3_68}
Mar{\'i}a Navas-Loro, V{\'i}ctor Rodr{\'i}guez-Doncel, Idafen Santana-Perez,
  and Alberto S{\'a}nchez. 2017.
\newblock Spanish corpus for sentiment analysis towards brands.
\newblock In \emph{Speech and Computer}, pages 680--689, Cham. Springer
  International Publishing.

\bibitem[{Nie et~al.(2020)Nie, Williams, Dinan, Bansal, Weston, and
  Kiela}]{nie-etal-2020-adversarial}
Yixin Nie, Adina Williams, Emily Dinan, Mohit Bansal, Jason Weston, and Douwe
  Kiela. 2020.
\newblock Adversarial {NLI}: A new benchmark for natural language
  understanding.
\newblock In \emph{Proceedings of the 58th Annual Meeting of the Association
  for Computational Linguistics}. Association for Computational Linguistics.

\bibitem[{Pang and Lee(2004)}]{pang-lee-2004-sentimental}
Bo~Pang and Lillian Lee. 2004.
\newblock \href {https://doi.org/10.3115/1218955.1218990} {A sentimental
  education: Sentiment analysis using subjectivity summarization based on
  minimum cuts}.
\newblock In \emph{Proceedings of the 42nd Annual Meeting of the Association
  for Computational Linguistics ({ACL}-04)}, pages 271--278, Barcelona, Spain.

\bibitem[{Pedregosa et~al.(2011)Pedregosa, Varoquaux, Gramfort, Michel,
  Thirion, Grisel, Blondel, Prettenhofer, Weiss, Dubourg, Vanderplas, Passos,
  Cournapeau, Brucher, Perrot, and Duchesnay}]{scikit-learn}
F.~Pedregosa, G.~Varoquaux, A.~Gramfort, V.~Michel, B.~Thirion, O.~Grisel,
  M.~Blondel, P.~Prettenhofer, R.~Weiss, V.~Dubourg, J.~Vanderplas, A.~Passos,
  D.~Cournapeau, M.~Brucher, M.~Perrot, and E.~Duchesnay. 2011.
\newblock Scikit-learn: Machine learning in {P}ython.
\newblock \emph{Journal of Machine Learning Research}, 12:2825--2830.

\bibitem[{Perez et~al.(2021)Perez, Kiela, and Cho}]{perez2021true}
Ethan Perez, Douwe Kiela, and Kyunghyun Cho. 2021.
\newblock \href {https://openreview.net/forum?id=ShnM-rRh4T} {True few-shot
  learning with language models}.
\newblock In \emph{Advances in Neural Information Processing Systems}.

\bibitem[{Petroni et~al.(2019)Petroni, Rockt{\"a}schel, Riedel, Lewis, Bakhtin,
  Wu, and Miller}]{petroni-etal-2019-language}
Fabio Petroni, Tim Rockt{\"a}schel, Sebastian Riedel, Patrick Lewis, Anton
  Bakhtin, Yuxiang Wu, and Alexander Miller. 2019.
\newblock \href {https://doi.org/10.18653/v1/D19-1250} {Language models as
  knowledge bases?}
\newblock In \emph{Proceedings of the 2019 Conference on Empirical Methods in
  Natural Language Processing and the 9th International Joint Conference on
  Natural Language Processing (EMNLP-IJCNLP)}, pages 2463--2473, Hong Kong,
  China. Association for Computational Linguistics.

\bibitem[{Reimers and Gurevych(2019)}]{reimers-2019-sentence-bert}
Nils Reimers and Iryna Gurevych. 2019.
\newblock \href {http://arxiv.org/abs/1908.10084} {Sentence-bert: Sentence
  embeddings using siamese bert-networks}.
\newblock In \emph{Proceedings of the 2019 Conference on Empirical Methods in
  Natural Language Processing}. Association for Computational Linguistics.

\bibitem[{Scherer and Wallbott(1994)}]{Scherer1994EvidenceFU}
K.~Scherer and H.~G. Wallbott. 1994.
\newblock Evidence for universality and cultural variation of differential
  emotion response patterning.
\newblock \emph{Journal of personality and social psychology}, 66 2:310--28.

\bibitem[{Schick and Sch{\"u}tze(2021)}]{schick-schutze-2021-exploiting}
Timo Schick and Hinrich Sch{\"u}tze. 2021.
\newblock \href {https://www.aclweb.org/anthology/2021.eacl-main.20}
  {Exploiting cloze-questions for few-shot text classification and natural
  language inference}.
\newblock In \emph{Proceedings of the 16th Conference of the European Chapter
  of the Association for Computational Linguistics: Main Volume}, pages
  255--269, Online. Association for Computational Linguistics.

\bibitem[{Snell et~al.(2017)Snell, Swersky, and Zemel}]{snell2017prototypical}
Jake Snell, Kevin Swersky, and Richard Zemel. 2017.
\newblock Prototypical networks for few-shot learning.
\newblock \emph{Advances in neural information processing systems}, 30.

\bibitem[{Song et~al.(2020)Song, Tan, Qin, Lu, and Liu}]{NEURIPS2020_c3a690be}
Kaitao Song, Xu~Tan, Tao Qin, Jianfeng Lu, and Tie-Yan Liu. 2020.
\newblock \href
  {https://proceedings.neurips.cc/paper/2020/file/c3a690be93aa602ee2dc0ccab5b7b67e-Paper.pdf}
  {Mpnet: Masked and permuted pre-training for language understanding}.
\newblock In \emph{Advances in Neural Information Processing Systems},
  volume~33, pages 16857--16867. Curran Associates, Inc.

\bibitem[{Troiano et~al.(2019)Troiano, Pad{\'o}, and
  Klinger}]{troiano-etal-2019-crowdsourcing}
Enrica Troiano, Sebastian Pad{\'o}, and Roman Klinger. 2019.
\newblock \href {https://doi.org/10.18653/v1/P19-1391} {Crowdsourcing and
  validating event-focused emotion corpora for {G}erman and {E}nglish}.
\newblock In \emph{Proceedings of the 57th Annual Meeting of the Association
  for Computational Linguistics}, pages 4005--4011, Florence, Italy.
  Association for Computational Linguistics.

\bibitem[{Vaswani et~al.(2017)Vaswani, Shazeer, Parmar, Uszkoreit, Jones,
  Gomez, Kaiser, and Polosukhin}]{NIPS2017_3f5ee243}
Ashish Vaswani, Noam Shazeer, Niki Parmar, Jakob Uszkoreit, Llion Jones,
  Aidan~N Gomez, \L~ukasz Kaiser, and Illia Polosukhin. 2017.
\newblock \href
  {https://proceedings.neurips.cc/paper/2017/file/3f5ee243547dee91fbd053c1c4a845aa-Paper.pdf}
  {Attention is all you need}.
\newblock In \emph{Advances in Neural Information Processing Systems},
  volume~30. Curran Associates, Inc.

\bibitem[{Vilares and
  G{\'o}mez-Rodr{\'\i}guez(2019)}]{vilares-gomez-rodriguez-2019-head}
David Vilares and Carlos G{\'o}mez-Rodr{\'\i}guez. 2019.
\newblock \href {https://doi.org/10.18653/v1/P19-1092} {{HEAD}-{QA}: A
  healthcare dataset for complex reasoning}.
\newblock In \emph{Proceedings of the 57th Annual Meeting of the Association
  for Computational Linguistics}, pages 960--966, Florence, Italy. Association
  for Computational Linguistics.

\bibitem[{Wang et~al.(2021)Wang, Fang, Khabsa, Mao, and
  Ma}]{Wang2021EntailmentAF}
Sinong Wang, Han Fang, Madian Khabsa, Hanzi Mao, and Hao Ma. 2021.
\newblock Entailment as few-shot learner.
\newblock \emph{ArXiv}, abs/2104.14690.

\bibitem[{Warstadt et~al.(2019)Warstadt, Singh, and
  Bowman}]{warstadt-etal-2019-neural}
Alex Warstadt, Amanpreet Singh, and Samuel~R. Bowman. 2019.
\newblock \href {https://doi.org/10.1162/tacl_a_00290} {Neural network
  acceptability judgments}.
\newblock \emph{Transactions of the Association for Computational Linguistics},
  7:625--641.

\bibitem[{Williams et~al.(2018)Williams, Nangia, and Bowman}]{N18-1101}
Adina Williams, Nikita Nangia, and Samuel Bowman. 2018.
\newblock \href {http://aclweb.org/anthology/N18-1101} {A broad-coverage
  challenge corpus for sentence understanding through inference}.
\newblock In \emph{Proceedings of the 2018 Conference of the North American
  Chapter of the Association for Computational Linguistics: Human Language
  Technologies, Volume 1 (Long Papers)}, pages 1112--1122. Association for
  Computational Linguistics.

\bibitem[{Wolf et~al.(2020)Wolf, Debut, Sanh, Chaumond, Delangue, Moi, Cistac,
  Rault, Louf, Funtowicz, Davison, Shleifer, von Platen, Ma, Jernite, Plu, Xu,
  Le~Scao, Gugger, Drame, Lhoest, and Rush}]{wolf-etal-2020-transformers}
Thomas Wolf, Lysandre Debut, Victor Sanh, Julien Chaumond, Clement Delangue,
  Anthony Moi, Pierric Cistac, Tim Rault, Remi Louf, Morgan Funtowicz, Joe
  Davison, Sam Shleifer, Patrick von Platen, Clara Ma, Yacine Jernite, Julien
  Plu, Canwen Xu, Teven Le~Scao, Sylvain Gugger, Mariama Drame, Quentin Lhoest,
  and Alexander Rush. 2020.
\newblock \href {https://doi.org/10.18653/v1/2020.emnlp-demos.6} {Transformers:
  State-of-the-art natural language processing}.
\newblock In \emph{Proceedings of the 2020 Conference on Empirical Methods in
  Natural Language Processing: System Demonstrations}, pages 38--45, Online.
  Association for Computational Linguistics.

\bibitem[{Yin et~al.(2019)Yin, Hay, and Roth}]{yin-etal-2019-benchmarking}
Wenpeng Yin, Jamaal Hay, and Dan Roth. 2019.
\newblock \href {https://doi.org/10.18653/v1/D19-1404} {Benchmarking zero-shot
  text classification: Datasets, evaluation and entailment approach}.
\newblock In \emph{Proceedings of the 2019 Conference on Empirical Methods in
  Natural Language Processing and the 9th International Joint Conference on
  Natural Language Processing (EMNLP-IJCNLP)}, pages 3914--3923, Hong Kong,
  China. Association for Computational Linguistics.

\bibitem[{Yin et~al.(2020)Yin, Rajani, Radev, Socher, and
  Xiong}]{yin-etal-2020-universal}
Wenpeng Yin, Nazneen~Fatema Rajani, Dragomir Radev, Richard Socher, and Caiming
  Xiong. 2020.
\newblock \href {https://doi.org/10.18653/v1/2020.emnlp-main.660} {Universal
  natural language processing with limited annotations: Try few-shot textual
  entailment as a start}.
\newblock In \emph{Proceedings of the 2020 Conference on Empirical Methods in
  Natural Language Processing (EMNLP)}, pages 8229--8239, Online. Association
  for Computational Linguistics.

\bibitem[{Zhang et~al.(2015)Zhang, Zhao, and LeCun}]{NIPS2015_250cf8b5}
Xiang Zhang, Junbo Zhao, and Yann LeCun. 2015.
\newblock \href
  {https://proceedings.neurips.cc/paper/2015/file/250cf8b51c773f3f8dc8b4be867a9a02-Paper.pdf}
  {Character-level convolutional networks for text classification}.
\newblock In \emph{Advances in Neural Information Processing Systems},
  volume~28. Curran Associates, Inc.

\end{thebibliography}
\bibliographystyle{acl_natbib}

% \newpage
\clearpage
\appendix

\begin{table*}[t]
\centering
\resizebox{\textwidth}{!}{%
\begin{tabular}{lrlllllllllll}
\toprule
language && \multicolumn{4}{c}{German} & \multicolumn{3}{c}{English} & \multicolumn{3}{c}{Spanish} \\
                 name &    n &                \gnad &       \amazonreviews &             \deisear &              \sbtenk &       \amazonreviews &             \semeval &             \unified &       \amazonreviews &              \headqa &               \sab s &          Mean \\
\midrule
               random &    0 &                 11.1 &                 20.0 &                 14.3 &                 33.3 &                 20.0 &                 33.3 &                 10.0 &                 20.0 &                 16.7 &                 33.3 &          21.2 \\
             FastText &    0 &         17.3$_{1.0}$ &         15.4$_{0.5}$ &         22.2$_{2.1}$ &         31.5$_{1.5}$ &         18.6$_{0.5}$ &         43.8$_{0.4}$ &         11.8$_{0.3}$ &         19.7$_{0.5}$ &         45.0$_{0.9}$ &  \newbftab{35.0}$_{2.2}$ &  26.0$_{1.2}$ \\
      xlm-roberta  &    0 &  \newbftab{37.8}$_{1.1}$ &  \newbftab{28.4}$_{0.7}$ &  \newbftab{43.1}$_{2.7}$ &  \newbftab{46.6}$_{1.3}$ &  \newbftab{35.4}$_{0.7}$ &  \newbftab{50.5}$_{0.4}$ &  \newbftab{21.3}$_{0.3}$ &  \newbftab{32.8}$_{0.6}$ &  \newbftab{50.6}$_{0.9}$ &         31.6$_{2.0}$ &  37.8$_{1.3}$ \\
\midrule
             Char-SVM &    8 &         56.1$_{2.8}$ &         30.5$_{2.2}$ &         29.4$_{1.6}$ &         45.4$_{2.5}$ &         30.0$_{1.6}$ &         33.6$_{1.1}$ &         12.2$_{1.1}$ &         30.8$_{1.2}$ &         36.3$_{2.6}$ &         50.6$_{5.3}$ &  35.5$_{2.5}$ \\
      xlm-roberta (FT) &    8 &  \underline{66.3}$_{3.7}$ &  \newbftab{45.1}$_{0.9}$ &  \newbftab{56.6}$_{2.1}$ &  \newbftab{55.9}$_{2.6}$ &  \underline{45.2}$_{1.2}$ &  \newbftab{55.7}$_{3.8}$ &  \newbftab{25.4}$_{0.7}$ &  \newbftab{42.5}$_{1.1}$ &  \newbftab{55.0}$_{2.3}$ &  \underline{58.1}$_{5.2}$ &  50.6$_{2.8}$ \\
      xlm-roberta (LT) &    8 &  \underline{64.6}$_{1.2}$ &         42.1$_{1.5}$ &         50.6$_{2.4}$ &         50.2$_{1.8}$ &         41.7$_{2.0}$ &         46.5$_{2.7}$ &         23.0$_{0.4}$ &         40.4$_{1.3}$ &  \underline{53.7}$_{2.9}$ &  \underline{52.2}$_{4.8}$ &  46.5$_{2.4}$ \\
 xlm-roberta (LT-DIST) &    8 &  \newbftab{67.0}$_{3.2}$ &  \underline{44.3}$_{0.8}$ &  \underline{53.2}$_{3.0}$ &  \underline{55.8}$_{2.0}$ &  \newbftab{45.4}$_{1.6}$ &  \underline{53.1}$_{3.3}$ &  \underline{25.3}$_{0.6}$ &  \underline{41.7}$_{1.4}$ &  \underline{54.6}$_{2.3}$ &  \newbftab{59.4}$_{4.2}$ &  50.0$_{2.5}$ \\
\midrule
             Char-SVM &   64 &         77.3$_{0.8}$ &         41.4$_{0.8}$ &         48.1$_{2.9}$ &         51.5$_{0.7}$ &         43.5$_{0.8}$ &         39.0$_{0.8}$ &         17.3$_{0.4}$ &         40.4$_{1.0}$ &         52.3$_{0.8}$ &         54.7$_{0.9}$ &  46.6$_{1.2}$ \\
      xlm-roberta (FT) &   64 &  \newbftab{79.7}$_{0.7}$ &  \newbftab{51.5}$_{1.0}$ &  \newbftab{67.7}$_{0.9}$ &  \newbftab{63.0}$_{0.9}$ &  \newbftab{53.1}$_{1.9}$ &  \newbftab{61.0}$_{1.6}$ &  \newbftab{28.1}$_{0.2}$ &  \newbftab{49.4}$_{0.3}$ &  \newbftab{60.5}$_{1.0}$ &  \underline{64.9}$_{1.8}$ &  57.9$_{1.2}$ \\
      xlm-roberta (LT) &   64 &         76.9$_{0.6}$ &         48.4$_{0.6}$ &         62.6$_{0.9}$ &         59.1$_{0.6}$ &         49.1$_{1.6}$ &         54.2$_{1.9}$ &         26.9$_{0.7}$ &         48.7$_{0.4}$ &  \underline{59.3}$_{0.8}$ &  \underline{61.8}$_{3.1}$ &  54.7$_{1.4}$ \\
 xlm-roberta (LT-DIST) &   64 &  \underline{78.9}$_{0.5}$ &  \underline{50.0}$_{1.1}$ &         64.7$_{0.3}$ &  \underline{62.5}$_{0.9}$ &  \underline{51.7}$_{1.3}$ &  \underline{59.5}$_{1.0}$ &  \underline{27.6}$_{0.4}$ &  \underline{48.9}$_{0.7}$ &  \underline{59.3}$_{0.9}$ &  \newbftab{65.4}$_{1.8}$ &  56.9$_{1.0}$ \\
\midrule
             Char-SVM &  512 &  \underline{85.0}$_{0.3}$ &         48.2$_{0.5}$ &         48.1$_{2.9}$ &         59.0$_{0.4}$ &         50.4$_{0.4}$ &         46.0$_{0.5}$ &         23.0$_{0.4}$ &         46.4$_{0.9}$ &         64.7$_{0.4}$ &         63.8$_{1.3}$ &  53.5$_{1.1}$ \\
      xlm-roberta (FT) &  512 &  \newbftab{85.4}$_{0.6}$ &  \newbftab{57.2}$_{0.7}$ &  \newbftab{67.8}$_{1.2}$ &  \newbftab{68.6}$_{0.9}$ &  \newbftab{58.8}$_{0.4}$ &  \newbftab{64.7}$_{0.7}$ &  \newbftab{32.1}$_{0.3}$ &  \newbftab{53.3}$_{0.6}$ &  \newbftab{68.8}$_{0.5}$ &  \newbftab{69.7}$_{0.5}$ &  62.6$_{0.7}$ \\
      xlm-roberta (LT) &  512 &         80.8$_{0.6}$ &         52.5$_{0.7}$ &         62.6$_{0.8}$ &         63.3$_{0.9}$ &         54.3$_{0.3}$ &         60.6$_{0.7}$ &         28.9$_{0.4}$ &         51.4$_{0.4}$ &         62.9$_{0.3}$ &         66.8$_{0.4}$ &  58.4$_{0.6}$ \\
 xlm-roberta (LT-DIST) &  512 &         80.7$_{0.4}$ &         54.1$_{0.3}$ &         64.6$_{0.2}$ &         66.0$_{1.3}$ &         55.6$_{0.3}$ &         62.9$_{1.0}$ &         30.5$_{0.4}$ &         52.4$_{0.2}$ &         63.1$_{0.4}$ &         68.7$_{0.6}$ &  59.9$_{0.6}$ \\
\bottomrule
\end{tabular}
}
\caption{Multi-lingual results for Siamese models based on \href{https://huggingface.co/sentence-transformers/paraphrase-multilingual-mpnet-base-v2}{paraphrase-multilingual-mpnet-base-v2}, comparing fine-tuning (FT), label tuning (LT) and label tuning with distillation (LT-DIST). Results are grouped by the number of training examples (n).  \underline{Underlined} results are significant. \textbf{Bold} font indicates maxima.}
\label{tab:results_sft_ml}
\end{table*}

\section{Unified Emotions}

Unified Emotions is a meta-dataset comprised of the following datasets: \dailydialog \citep{li2017dailydialog}, \crowdflower \citep{crowdflower2016}, \tec \citep{mohammad-2012-emotional},
\talesemotion \citep{alm2005emotions,alm2005perceptions,alm2008affect}, \isear \citep{Scherer1994EvidenceFU},
\emoint \citep{mohammad2017stance}, \electoraltweets \citep{mohammad2015sentiment}, \groundedemotions \citep{Liu17} and \emotioncause \citep{ghazi2016detecting}.

\section{Hypotheses}
\label{sec:entail_patterns}

Table \ref{tab:entail_patterns} lists all the hypothesis patterns used in our experiments.
 
% https://docs.google.com/document/d/1XyjuPBjenDvCFJJCw8uLWTUhjzLaW3i-Qak-6gxXPLw
\begin{table*}
\resizebox{\textwidth}{!}{%
\begin{tabular}{p{2.5cm}p{2.5cm}p{1.0cm}p{10cm}}
\toprule
dataset    & type  & lang. & pattern \\
\midrule
\unified & Emotions &  en & This person feels \{anger, disgust, feat, guilt, joy, love, sadness, shame, surprise\}.\\
         &&& This person doesn't feel any particular emotion.\\
\deisear &          & de & Diese Person empfindet \{Schuld, Wut, Ekel, Angst, Freude, Scham, Traurigkeit\}.\\ 
\midrule
\agnews  & Topic & en& It is \{business, science, sports, world\} news.\\
\gnad  &       & de & Das ist ein Artikel aus der Rubrik \{Web, Panorama, International, Wirtschaft, Sport, Inland, Etat, Wissenschaft, Kultur\}.\\
\headqa  &       & es & Está relacionado con la \{medicina, enfermería, química, biología, psicología, farmacología\}.\\
\midrule
\yahoo &  & en & It is related with \{business \& finance,computers \& internet, education \& reference, entertainment \& music, family \& relationships, health, politics \& government, science \& mathematics, society \& culture, sports\}.\\
\amazonreviews & Review & en & This product is \{terrible, bad, okay, good, excellent\}.\\
               &      &  de & Dieses Produkt ist \{furchtbar, schlecht, ok, gut, exzellent\}.\\
               &      & es  & Este producto es \{terrible, mal, regular, bien, excelente\}.\\
\imdb, \yelp (2) &  & en &  It was \{terrible, great\}.\\
\yelp (5) & & & It was \{terrible, bad, okay, good, great\}.\\
\midrule
\semeval{} & Sentiment & en & This person expresses a \{negative, neutral, positive\} feeling.\\
\sbtenk{} &            & de & Diese Person drückt ein \{negativ, neutral, positiv\}es Gefühl aus.\\
\sab{}    &            & es & Esta persona expresa un sentimiento \{negativo, neutro, positivo\}.\\
\midrule
\cola & Acceptability & en & It is \{correct, incorrect\}.\\
\subj & Subjectivity & en & It is \{objective, subjective\}.\\
\trec & Question Type & en &It is \{expression, description, entity, human, location, number\}.\\
\bottomrule
\end{tabular}
}
\caption{Hypotheses patterns used.}
\label{tab:entail_patterns}
\end{table*}

\section{Paraphrase datasets}

\texttt{paraphrase-mpnet-base-v2} has been trained on these datasets: 
SNLI,
MNLI,
sentence-compression,
SimpleWiki,
altlex,
msmarco-triplets,
quora\_duplicates,
coco\_captions,
yahoo\_answers\_title\_question,
S2ORC\_citation\_pairs,
stackexchange\_duplicate\_questions and
wiki-atomic-edits.
Details on these dataset are provided \href{https://www.sbert.net/examples/training/paraphrases/README.html#datasets}{here}.

\section{Hyperparameters}
\label{sec:hparams}
For the label tuning experiments we used the following hyper-parameters:

\begin{itemize}
    \itemsep0em
    \item learning rate $\in \{0.01, 0.1\}$
    \item number of epochs $\in \{1000, 2000\}$
    \item regularizer coefficient $\in \{0.01, 0.1\}$
    \item dropout rate $\in \{0.01, 0.1\}$
\end{itemize}

\section{Additional Analysis}
\label{sec:analisis} 

The following table shows the F1-score breakdown by hypothesis length.
One could think that the CA model performs better for longer hypothesis but this cannot be observed.
Potentially because all hypotheses are relatively short.

\begin{table}[H]
\centering
% \resizebox{\columnwidth}{!}{%
\begin{tabular}{lrrr}
\toprule
name &   3-5 &   5-7 &    $>$7 \\
\midrule
  SN &  42.2 &  32.9 &  30.3 \\
  CA &  41.4 &  30.1 &  25.2 \\
\bottomrule
\end{tabular}
% }
\caption{Average macro F1 score by length of the reference hypothesis, averaged over all test sets for $n{=}0$.}
% \label{tab:a_hypo_length}
\end{table}

For completeness, we also add similar breakdowns by task type, label set size, and language.
None of them indicate an effect on the difference between SN and CA model performance.

\begin{table}[H]
\centering
% \resizebox{\columnwidth}{!}{%
\begin{tabular}{lrrr}
\toprule
name &   2-3 &   4-6 &   $>$6 \\
\midrule
  SN &  51.1 &  36.7 &  34.7 \\
  CA &  52.1 &  32.0 &  31.2 \\
\bottomrule
\end{tabular}
% }
\caption{Average macro F1 score by label set size, averaged over all test sets for $n{=}0$.}
\end{table}

\begin{table}[H]
% \centering
\resizebox{\columnwidth}{!}{%
\begin{tabular}{lrrrrr}
\toprule
name &  emotions &  other &  reviews &  sentiment &  topic \\
\midrule
  SN &      21.8 &   40.4 &     46.4 &       39.0 &   48.3 \\
  CA &      22.2 &   34.7 &     47.8 &       33.7 &   44.4 \\
\bottomrule
\end{tabular}
}
\caption{Average macro F1 score by task, averaged over all test sets for $n{=}0$.}
\end{table}

\begin{table}[H]
\centering
% \resizebox{\columnwidth}{!}{%
\begin{tabular}{lrrr}
\toprule
name &    de &    en &    es \\
\midrule
  SN &  33.3 &  47.7 &  31.8 \\
  CA &  27.0 &  46.8 &  30.1 \\
\bottomrule
\end{tabular}
% }
\caption{Average macro F1 score by language, averaged over all test sets for $n{=}0$.}
\label{tab:a_hypo_length}
\end{table}

\section{Computing Requirements}
\label{sec:hardware}

All experiments were run on a system with an AMD Ryzen Threadripper 1950X CPU and a Nvidia GeForce GTX 1080 Ti GPU.
Most of the computing time was spent training the NLI models used in our experiments.
Training the CA models took approx. 20h while training the SN models took approx. 10h.

\section{NLI Training sets}
\label{sec:nli_train}
\begin{table}[H]
\centering
% \resizebox{\columnwidth}{!}{%
\begin{tabular}{lr}
\toprule
name & examples \\
\midrule
SNLI \cite{bowman-etal-2015-large} & 569,033 \\
MNLI \cite{N18-1101} & 412,349 \\
ANLI \cite{nie-etal-2020-adversarial} & 169,246 \\
XNLI \cite{conneau2018xnli} & 112,500 \\
\bottomrule
\end{tabular}
% }
\caption{Sizes of NLI traininig sets. SNLI, MNLI and ANLI are English only. XNLI contains 15 languages with 7,500 examples per language.}
\end{table}

\section{Multilingual Label Tuning Results}

\label{sec:results_sft_ml}

Table \ref{tab:results_sft_ml} multilingual results for label tuninig and fine-tuning.

\end{document}